\documentclass[letterpaper,12pt,titlepage]{article}
\usepackage[margin=1in]{geometry}
\usepackage{setspace}

\usepackage{cite}
\usepackage{amsmath,amssymb,amsfonts}
\usepackage{algorithmic}
\usepackage{graphicx}
\usepackage{textcomp}
\usepackage{xcolor}
\usepackage{url}
\usepackage{float}
\usepackage{placeins}
\usepackage{hyperref}

\usepackage{amsthm}
\usepackage[shortlabels]{enumitem}
\usepackage{caption}
\usepackage{subcaption}
\usepackage{array,multirow}

\theoremstyle{plain}

\theoremstyle{definition}

\title{Accurate Crop Yield Estimation of Blueberries using Deep Learning and Smart Drones}

\author{Hieu D. Nguyen$^1$ \and
Brandon McHenry$^2$ \and
Thanh Nguyen$^1$ \and
Harper Zappone$^2$ \and
Anthony Thompson$^2$ \and
Chau Tran$^2$ \and
Anthony Segrest$^2$ \and
Luke Tonon$^2$}

\date{$^1$ Department of Mathematics \\$^2$ Department of Computer Science \\Rowan University, Glassboro, NJ USA \\ November 16, 2024}

\begin{document}
\maketitle
\pagestyle{plain}

\begin{abstract}
We present an AI pipeline that involves using smart drones equipped with computer vision to obtain a more accurate fruit count and yield estimation of the number of blueberries in a field. The core components are two object-detection models based on the YOLO deep learning architecture: a Bush Model that is able to detect blueberry bushes from images captured at low altitudes and at different angles, and a Berry Model that can detect individual berries that are visible on a bush.  Together, both models allow for more accurate crop yield estimation by allowing intelligent control of the drone's position and camera to safely capture side-view images of bushes up close.  In addition to providing experimental results for our models, which show good accuracy in terms of precision and recall when captured images are cropped around the foreground center bush, we also describe how to deploy our models to map out blueberry fields using different sampling strategies, and discuss the challenges of annotating very small objects (blueberries) and difficulties in evaluating the effectiveness of our models.
\end{abstract}

%\begin{IEEEkeywords} blueberry crop yield, computer vision, smart drones, precision agriculture
%\end{IEEEkeywords}

\section{Introduction}
% -----------------------------------------------
Precision agriculture using AI and autonomous drones has been shown to be an effective approach in not only estimating yield for many different crops but also in detecting and managing weeds and diseases.  One popular specialty crop where accurate estimation of crop yield is important is the blueberry, where prediction early in the growing season is important in helping farmers make pricing decisions, hire a sufficient number of pickers, and inform their distributors of available supply before harvest time.  

One AI-based approach to estimating crop yield involves using deep learning models trained on images to detect fruits and vegetables; many such models have been developed for many of them, including blueberries, grapes, apples, and tomatoes. However, early models were trained on either simulated data [4] or close-up images of clusters of fruit (from hand-held cameras or mobile devices) and not the entire bush or vine [5-9]; thus, these models are inappropriate for industrial use where crop count and disease detection must be performed over a large field many acres in size.  For example, a one-acre blueberry field can contain up to 1000 bushes, which would make walking through such a field to capture a good sample of close-up images quite time-consuming; thus, a ``boots-on-the-ground" approach is impractical.  

On the other hand, using unmanned aerial vehicles (UAVs), in our case drones, to capture images of blueberry bushes from a farther distance presents a much more efficient approach to mapping large fields\footnote[1]{We note that ground autonomous vehicles (or robots) offer an alternative solution; however, we shall not discuss them and their trade-offs in this paper since this topic has been well addressed in the literature and doing so will set us too far adrift from the focus of our work.}.
In particular, we consider smart drones programmed with computer vision, i.e., drones that have an onboard mini-computer to process images captured by an onboard camera, to identify a blueberry bush and position itself to capture an optimal view of the bush that maximizes the number of visible berries on the bush. In addition, we make a distinction between a smart drone vs an autonomous drone, where the former is capable of making its own decisions during flight, e.g., by altering its path to avoid collision, as opposed to an autonomous drone where its actions, including its flight path, is pre-programmed before take-off.  For example, commercial drones that are equipped with collision-avoidance systems (based on technologies such as infrared, stereo vision, and LiDAR) and provide object-tracking capabilities would fall under our classification of a smart drone.

In this work, we present a pipeline of object-detection models based on the YOLO (You Only Look Once \cite{Redmon_2016_CVPR}) deep learning architecture to estimate the crop yield of a blueberry field by using \textit{smart} drones programmed with these models to accurately capture images of blueberry bushes and detect the number of harvestable berries on each bush (Figure \ref{fig:bb-bush} shows an example of a blueberry bush).  Our work is novel in its approach of detecting not only individual berries that are visible on an entire blueberry bush but also in detecting the bush itself in order to guarantee accurate drone position and image capture of the bush.  We distinguish such a smart mission from an autonomous mission as follows: in the latter, a drone is programmed before the start of the mission to fly to predetermined points of a blueberry field, say, along a row of bushes but keeping itself some fixed distance away from the row (and high enough to clear neighboring rows); moreover, its onboard camera can be programmed to capture images of these bushes, but from a fixed angle of view.  However, this does not always guarantee that the bushes will be fully captured by the drone's camera. For example, this situation may occur if the drone's position is blown off course due to wind or temporary loss of GPS. Thus, it is not possible for the drone to adjust its distance from the bushes or its camera's angle of view in real time to compensate for this.

In contrast, drones programmed with our Bush Model can fly a more intelligent mission where it is able to adjust its position and fly as close as possible alongside a bush (while maintaining a safe distance) in order to capture an optimal view of the bush so that berries on it appear as largest as possible (for more accurate detection).  This also allows for the implementation of various sampling strategies, e.g., random stratified sampling of bushes, without having to know the exact GPS location of each bush, something that would be required for autonomous missions.  On the other hand, coupled with real-time kinematic (RTK) positioning, our pipeline makes precision mapping of blueberry fields possible where the location of each bush can be geotagged using our Bush  Model. Berry count can then calculated for each bush post flight using our Berry Model.  Then given appropriate field data, which we describe in Section III, crop yield can be estimated more accurately than from current methods.

The rest of the paper is divided as follows.  In Section II, we discuss related works.  In Section III, we present our proposed pipeline for mapping a blueberry field to obtain a more precise estimation of crop yield. 
 In Section IV, we present experimental results for our models, which we hope provides baseline results for future researchers to compare their work against, and discuss challenges we faced in annotating tiny objects (blueberries) and how this impacted the effectiveness of our models.  

\section{Related Works}

Recent advances in computer vision, in particular object-detection models based on deep learning such as the YOLO architecture, have led to a proliferation of works in precision agriculture.  More specifically, many of these works present highly accurate models to perform fruit detection and yield estimation.  Since many articles have recently been published in this field (a 2020 survey article \cite{cropyield-survey} reviewed 30 articles that employed deep learning models),  we limit our discussion to works that involve either deep learning models for detecting blueberries or drone-based methods.

Works similar to ours include \cite{apple-faster-rcnn} where a pipeline for estimating crop yield of apples using drones and objection-detection models based on the Faster R-CNN and SSD-Mobilenet architectures is described; \cite{apple-yolo-improved, intelligent-fruit-detection} where improved YOLOv5 models are used to detect apples from aerial drone images; \cite{apple-yield-mapping} where a semantic segmentation approach is used to detect and count apples but trained on ground-based images; \cite{drone-tomato} where a YOLOv5 is trained to detect tomatoes from drone-based images; \cite{yield-precision-ag} where YOLO-based models are used to detect apples, oranges, and pumpkins; and \cite{treetop} where a YOLOv5 model is trained to detect black pine tree tops from UAV images.  

\begin{figure}[t!]
\centering
\includegraphics[width=0.48\columnwidth]{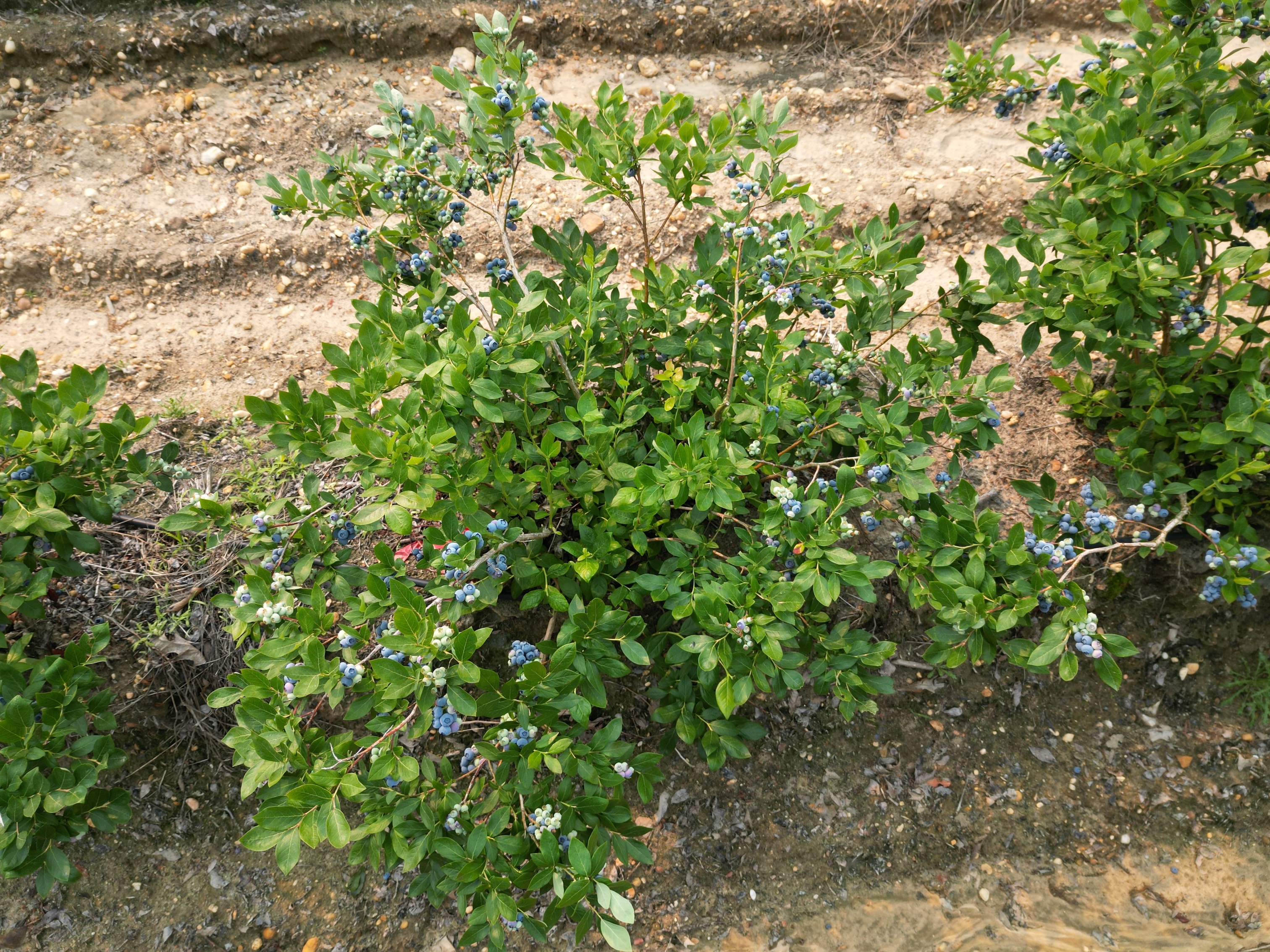}
\caption{Image of a blueberry bush}
\label{fig:bb-bush}
\end{figure}

We note that the aforementioned fruits and trees in the articles cited previously are somewhat large in comparison to blueberries, which, due to their size, are much more difficult to accurately annotate, especially in aerial drone images where the spatial resolution of the berries is poor due to their small size and distance from the drone (see Figure \ref{fig:bb-bush}).  Works related to fruits  that are comparable in size include \cite{cranberry} where a deep-learning model based on U-Net and trained on aerial images is described to segment and count cranberries for yield estimation and sun exposure; \cite{grape-cluster-yolo} where YOLO-based models are used to perform real-time tracking and counting of grape clusters; and \cite{grape-bunch-damage-yolo} where YOLO-based models are used to not only detect grape bunches but also assess their quality in terms of damage from lesions.

Works involving blueberries include \cite{blueberry-maturity-yield, blueberry-ripeness} where YOLOv3-v4 models are used 
YOLOv3-v4 to detect and estimate different stages of ripeness in blueberries (the latter in wild blueberries), but trained on images captured by hand-held cameras; \cite{blueberry-traits-segmentation} where a Mask R-CNN model is employed to segment individual blueberries in order to estimate fruit maturity;
\cite{blueberry-row-segmentation}
where a row detection segmentation model based on the U-Net architecture and trained on UAV images is described; and \cite{bush-detection}, where a bush-detection model is described, but trained on annotations of only the trunks of blueberry bushes and not the entire bush; on the other hand, our Bush Model is trained to detect the entire bush, which is necessary for accurate crop yield estimation.

There are very few works that are similar to ours where their computer vision models are validated on images and the results are compared against the actual fruit count per plant or tree. In particular, for our validation dataset, berries on 15 blueberry bushes were all hand-picked to obtain actual fruit counts (what we call, the "picked ground truth"). These works are few in number because actual or harvested crop yield is typically recorded by weight and not by the number of fruit. Among such works, most involve large fruit such as apples \cite{apple-fruit-actual-count} and mangos \cite{mango-fruit-actual-count, mango-fruit-segmentation-actual-count}, but also for almonds \cite{almond-fruit-count} and grapes \cite{grape-fruit-count}.  However, none of these works explicitly discuss the ratio of visual fruit count to actual fruit count (per bush) as we do in Section 6.4 to better understand the amount of occlusion.

\section{Pipeline}
\begin{figure}[tph]
\centering
\includegraphics[width=0.48\textwidth]{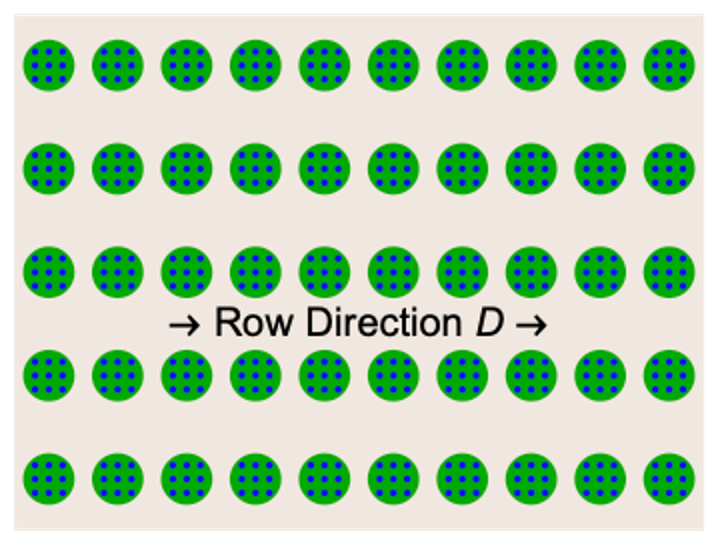}
\caption{Row direction of a field}
\label{fig:row-direction}
\end{figure}

 Let $F$ be a two-dimensional rectangular blueberry field of size $A$ (in acres) containing $C$ bushes.  We assume that GPS coordinates of the corners of $F$ and the direction $D$ of the rows of bushes in $F$ are known; moreover, we assume all rows run along the same direction (see Figure \ref{fig:row-direction}).  Let $Y$ denote the crop yield of $F$ (berries/acre).  

 Our pipeline for estimating $Y$ can be summarized as follows: we first use stratified sampling to fly a smart drone over random points of $F$ and capture images of bushes at these points using our Bush Detection model.  We then calculate the number of berries on these bushes using our Berry Detection model to estimate $Y$ using formula \ref{eq:crop-yield}.  Here the main steps of our pipeline depending on the sampling method.
\begin{enumerate}

\item Stratified Sampling: We partition $F$ into a grid of $M \times N$ non-overlapping square cells, denoted by $\{C_{mn}\}$. We present two stratified sampling strategies to sample bushes within each cell: point sampling and row sampling.  In point sampling, we select a single bush closest to a randomly chosen point inside each $C_{mn}$.  In row sampling, we sample a row of bushes inside $C_{mn}$.  See Figure \ref{fig:stratified-sampling}.

\begin{enumerate}
    \item Point Sampling: Select a random point, denoted by $p_{mn}$, inside each cell $C_{mn}$.
    \item Row Sampling: Select a random point $p_{mn}$ along an edge of $C_{mn}$ whose direction is perpendicular to $D$.
    \item Fly drone to each position $p_{mn}$ (at given altitude $h$).
\end{enumerate}

\begin{figure}[tph]
\centering
\includegraphics[width=0.48\textwidth]{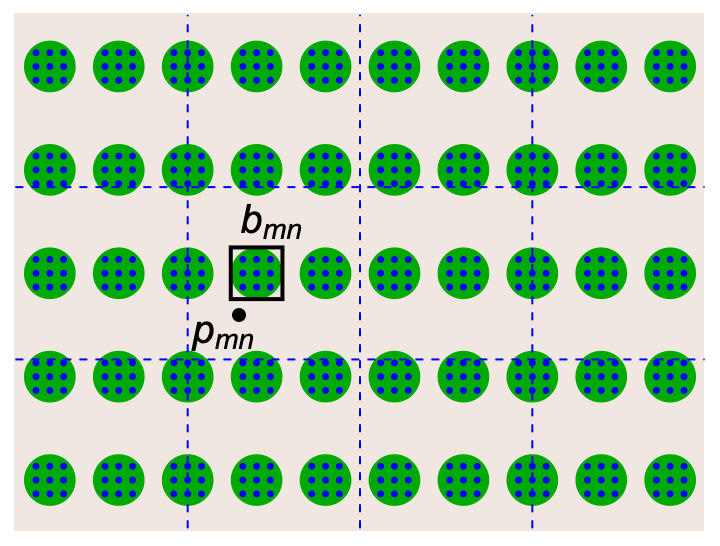}
\caption{Stratified sampling of a field}
\label{fig:stratified-sampling}
\end{figure}

\item Single Bush Detection
\begin{enumerate}
    \item At position $p_{mn}$, use drone camera (pointed down) to capture bushes in its angle of view and use our Bush Model to identify a bush closest to $p_{mn}$.  Denote by $b_{mn}$ the position of this closest bush, i.e., the center of the bounding box enclosing the bush as illustrated in Figure \ref{fig:stratified-sampling}).  Then apply object tracking using DeepSort to begin tracking the bush.
    \item Program drone to fly horizontally to position $b_{mn}$ (while maintaining altitude $h$) so that the drone is directly over the bush.
\end{enumerate}

\begin{figure}[tph]
\centering
\includegraphics[width=0.48\textwidth]{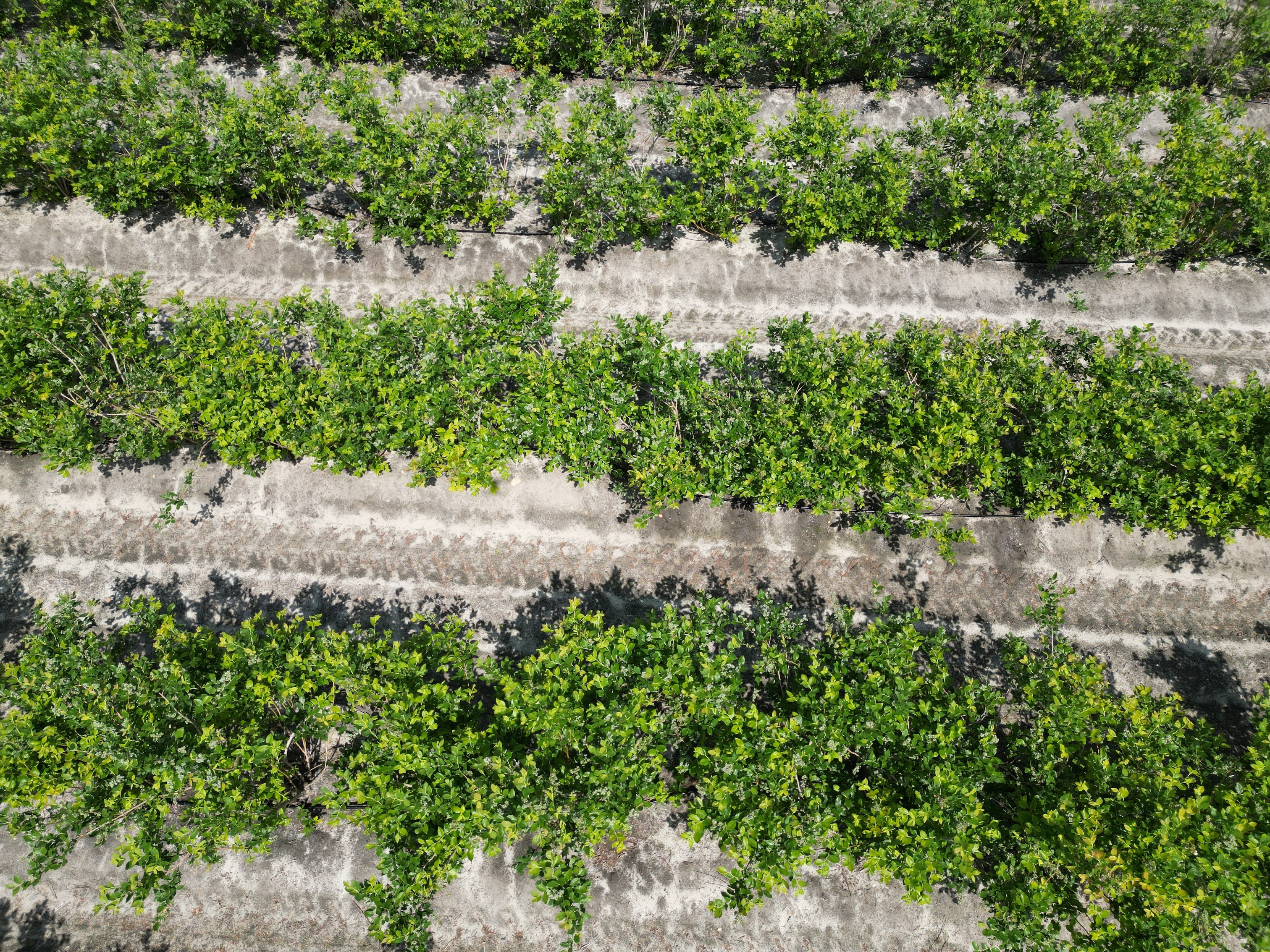}
\caption{Blueberry field from a birds-eye view}
\label{fig:bb-field-be}
\end{figure}

\begin{figure}[tph]
\centering
\includegraphics[width=0.48\textwidth]{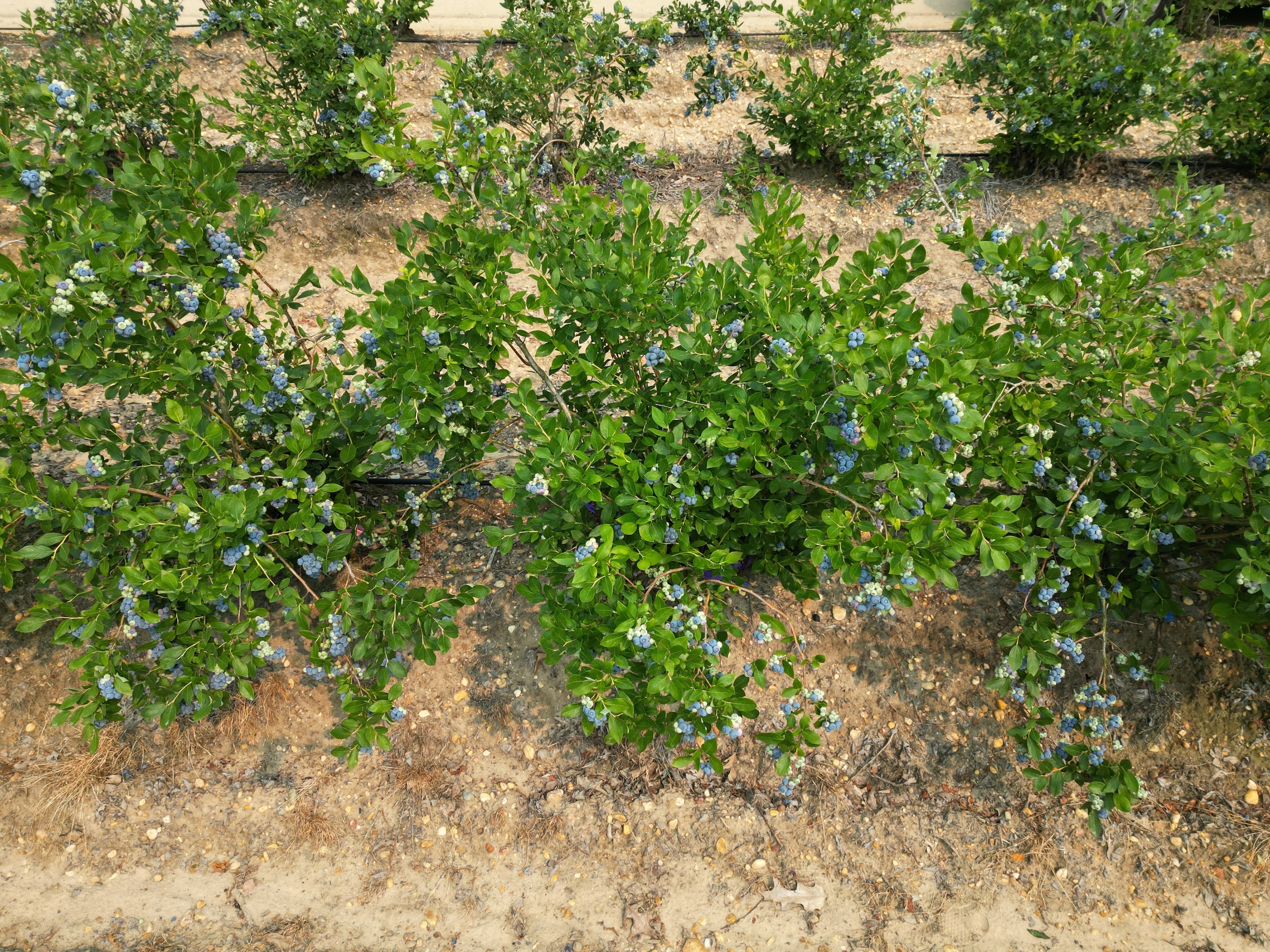}
\caption{Row of blueberry bushes from a angled-side view}
\label{fig:bb-row-sa}
\end{figure}

\item Bush Image Capture (Angled-Side View)
\begin{enumerate}
    \item Fly drone horizontally to one side of the bush (side chosen randomly to account for factors such as sun angle) in a direction perpendicular to $D$ (see Figure \ref{fig:row-direction}) so that it is distance $d$ away from $b_{mn}$ while simultaneously adjusting the camera angle to keep the bush in view using object tracking.  Further adjustments to either drone position or camera angle (or combination of both) can be made so that the entire bush is in view of camera (see Figure \ref{fig:bb-row-sa}).
    \item Point Sampling: Use drone camera to capture image of bush (at position $b_{mn}$) and record coordinates of bounding box predicted by Bush Model, denoted by $c_{mn}$.
    \item Row Sampling: Fly drone along $D$ and use camera to capture photos of bushes that appear in the center of its view and record the bounding box coordinates $c_{mn}$ of each captured bush until drone reaches opposite edge of the cell $C_{mn}$.  Apply object tracking to distinguish different bushes and adjust the drone's position to ensure photo capture of the entire bush.
\end{enumerate}
\item Berry Counting (Post-mission)
\begin{enumerate}
    \item Use bounding box coordinates $c_{mn}$ to determine the number of visible berries on the corresponding bush (from one side) captured by images in previous step.  We describe two approaches (we discuss their trade-offs in Section IV):
    \begin{enumerate}
        \item  Image Cropping: Crop each image to contain only the bush (with bounding box coordinates $c_{mn}$) and apply Berry Model on cropped image to obtain the number of visible berries on each bush.
        \item Bounding Box Filtering: Apply Berry Model on entire image and count only detections of berries that are inside the bush's bounding box to obtain the number of visible berries on each bush. 
        \end{enumerate}
    \item Calculate the mean number of berries obtained in the previous step (averaged over all bushes) and double this answer to obtain the mean number of berries per bush, denoted by $B$.
\end{enumerate}
\item Crop Yield Estimation: We calculate crop yield $Y$ as follows:
\begin{equation} \label{eq:crop-yield}
    Y=\alpha \cdot \frac{B\cdot C}{A}
\end{equation}
where $B$ is determined from the previous step, $C$ is the bush count, $A$ is the size of $F$, and $\alpha$ is a fixed proportionality constant, called the Picked-Visual Ratio (PVR), that describes the ratio of what we refer to as the \textit{picked ground truth} to the \textit{visual ground truth}:
\begin{equation}
\alpha=\frac{\textrm{Picked GT}}{\textrm{Visual GT}}
\end{equation}
The picked ground truth is defined to be the number of berries that are actually on a bush (or on a group of bushes) and verified by manually picking and counting all the berries on that bush.  On the other hand, the visual ground truth is defined to be the number of berries on the same bush that is visible from a side view and ideally given by $B$ if our Berry Model was perfect.  We assume $\alpha$ to be given and that it would be determined from historical data since $\alpha$ would be highly dependent on factors such as climate, soil, variety of blueberry, and height and angle of drone camera.  In Section VI we provide first estimates of $\alpha$ that were obtained from two validation datasets where all the berries on the foreground center bush of each image were picked by hand by our team to obtain the picked ground truth.
\end{enumerate}

\section{Datasets}
% -----------------------------------------------

\subsection{Data Collection}
% ---------------------
Our data consists of images (still photos and video frames) of blueberry bushes (highbush blueberry, \textit{Vaccinium corymbosum}) from two different varieties, Duke and Draper. These imagers were collected at outdoor blueberry farms in southern counties of New Jersey and captured using a combination of hand-held (including cellphone) cameras and drone cameras to create a more diverse dataset. Although our total collection consists of over a thousand such images, only a fraction of them were used to train our models due to limited resources and the time-consuming process of annotating these images, which we further discuss below. 
 
The following datasets were created to train and validate our Berry and Bush Models. 

\subsection{Berry Datasets}
% ---------------------

The Berry datasets in total consist of 95 annotated images that were either captured using drone cameras or handheld cameras:
\begin{enumerate}
    \item 35 aerial photos (drone)
    \item 60 ground photos (hand-held)
\end{enumerate}

These 95 images represent the total number that we have been able to completely annotate to date.  Although relatively few in number, these images contain well over 100,000 annotations of berries, which we discuss later in Section 4.4.  Thus, we believe that the size of our datasets at this point is sufficient to obtain reasonably accurate models as we demonstrate later.

\begin{figure}[ht!]
\begin{subfigure}[b]{0.48\columnwidth}
\centering
\includegraphics[width=0.89\textwidth]{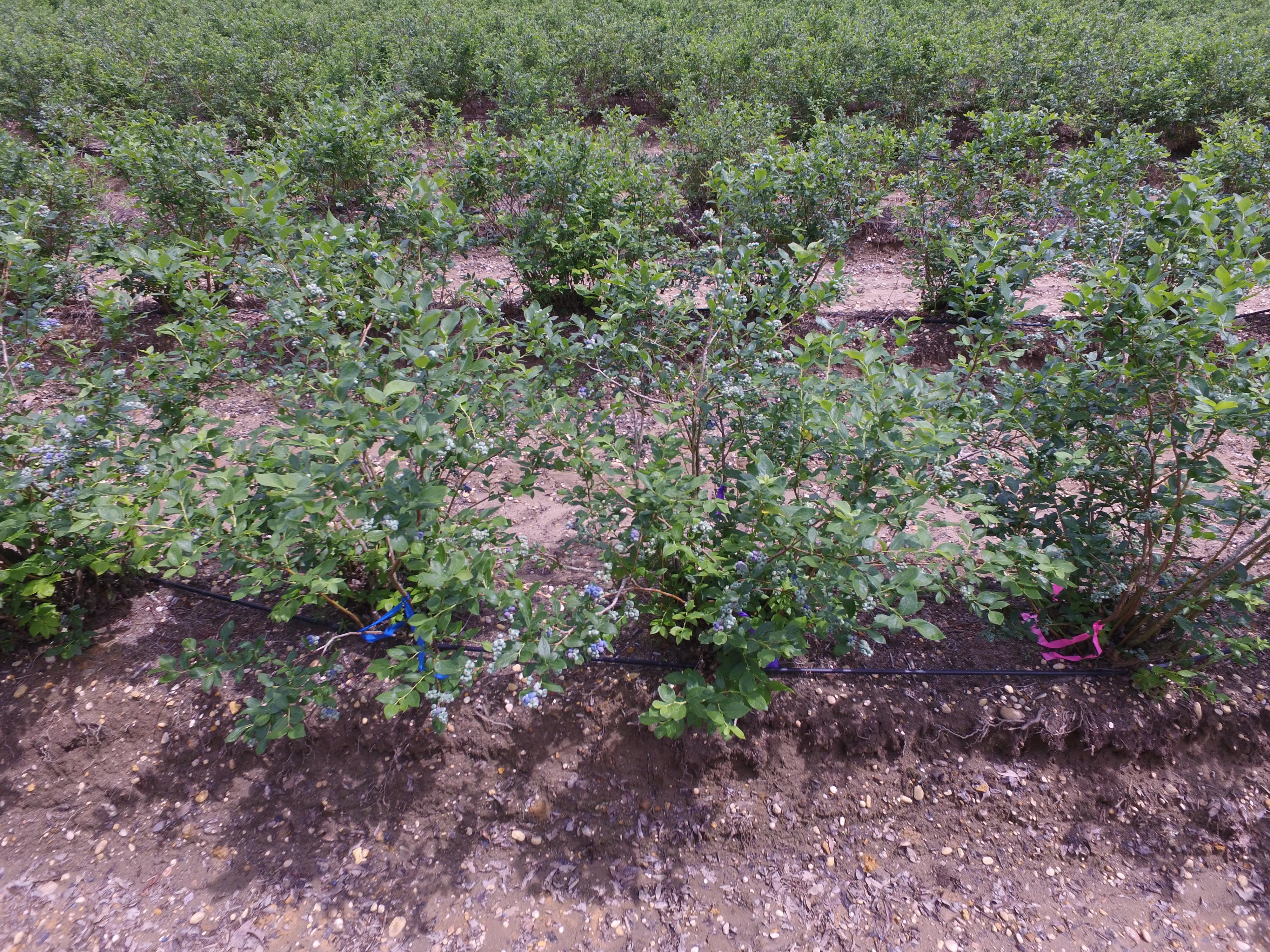}
\caption{DJI PHantom 3}
\label{fig:drone-dji}
\end{subfigure}%
~
\begin{subfigure}[b]{0.48\columnwidth}
\centering
\includegraphics[width=1\columnwidth]{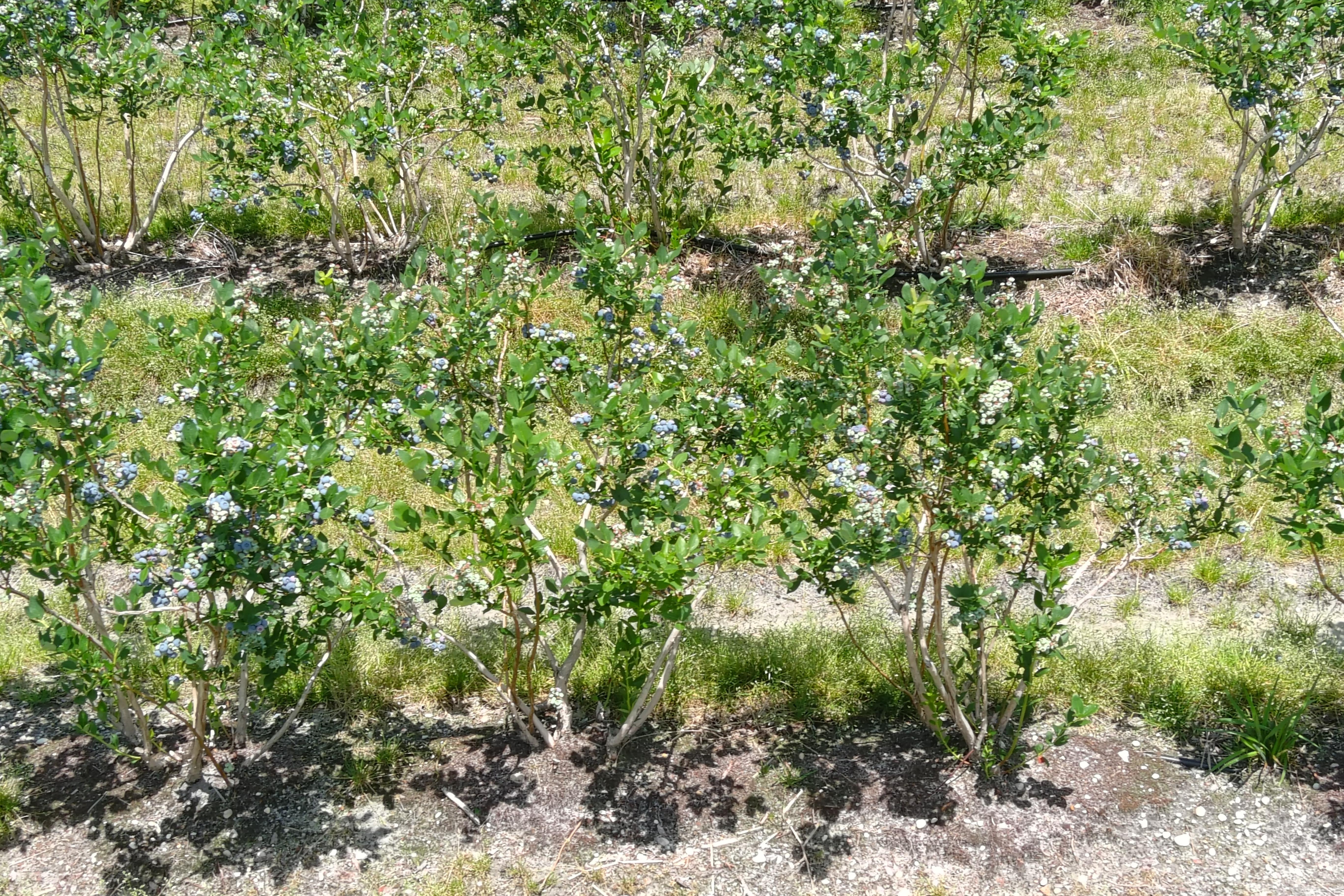}
\caption{Autel Evo II}
\label{fig:drone-evo}
\end{subfigure}
\caption{Sample drone images from Training Set.}
\end{figure}

\begin{figure}[ht!]
\begin{subfigure}[b]{0.48\columnwidth}
\centering
\includegraphics[width=1\textwidth]{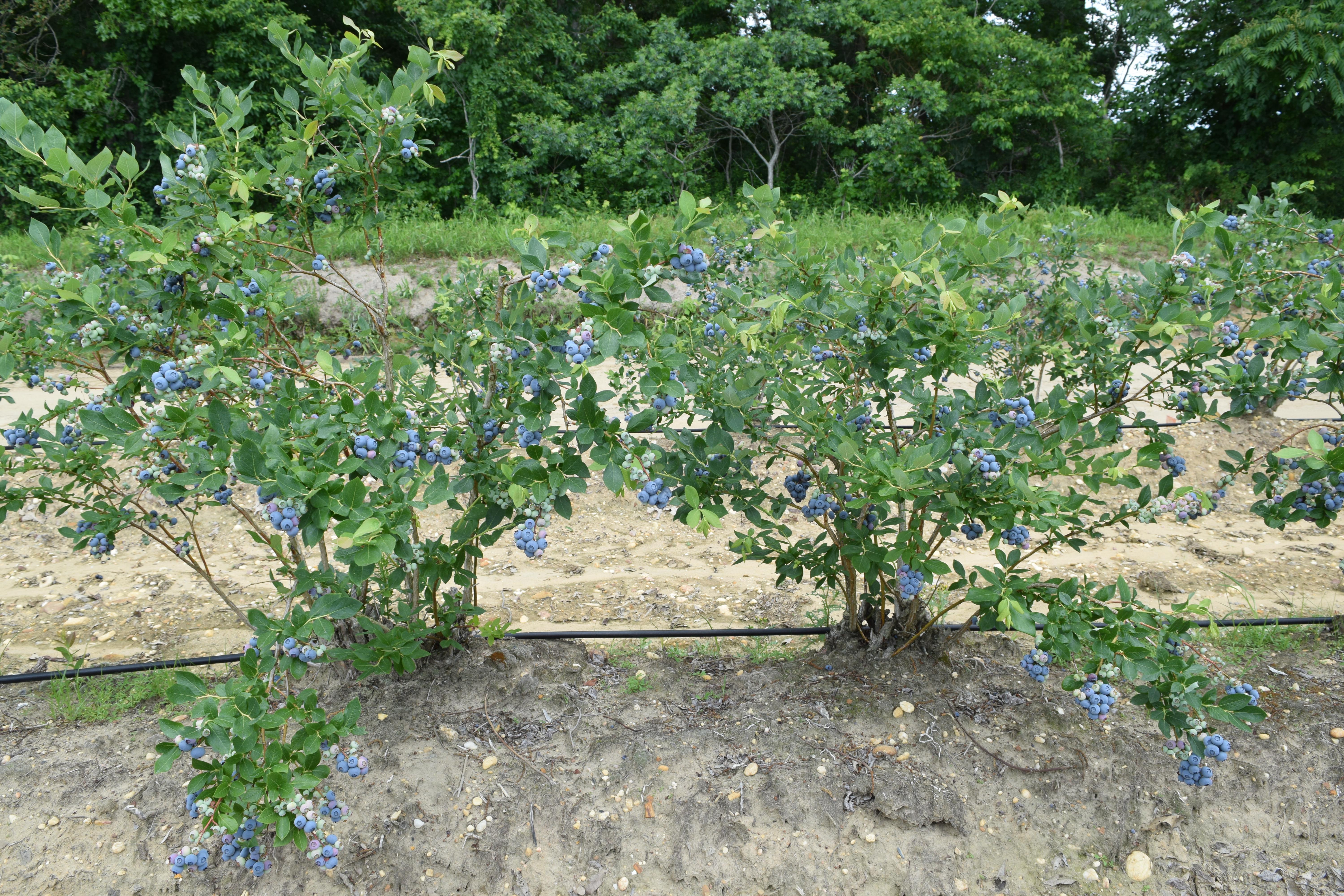}
\caption{Canon EOS Rebel}
\label{fig:handheld-rebel}
\end{subfigure}%
~
\begin{subfigure}[b]{0.48\columnwidth}
\centering
\includegraphics[width=.89\columnwidth]{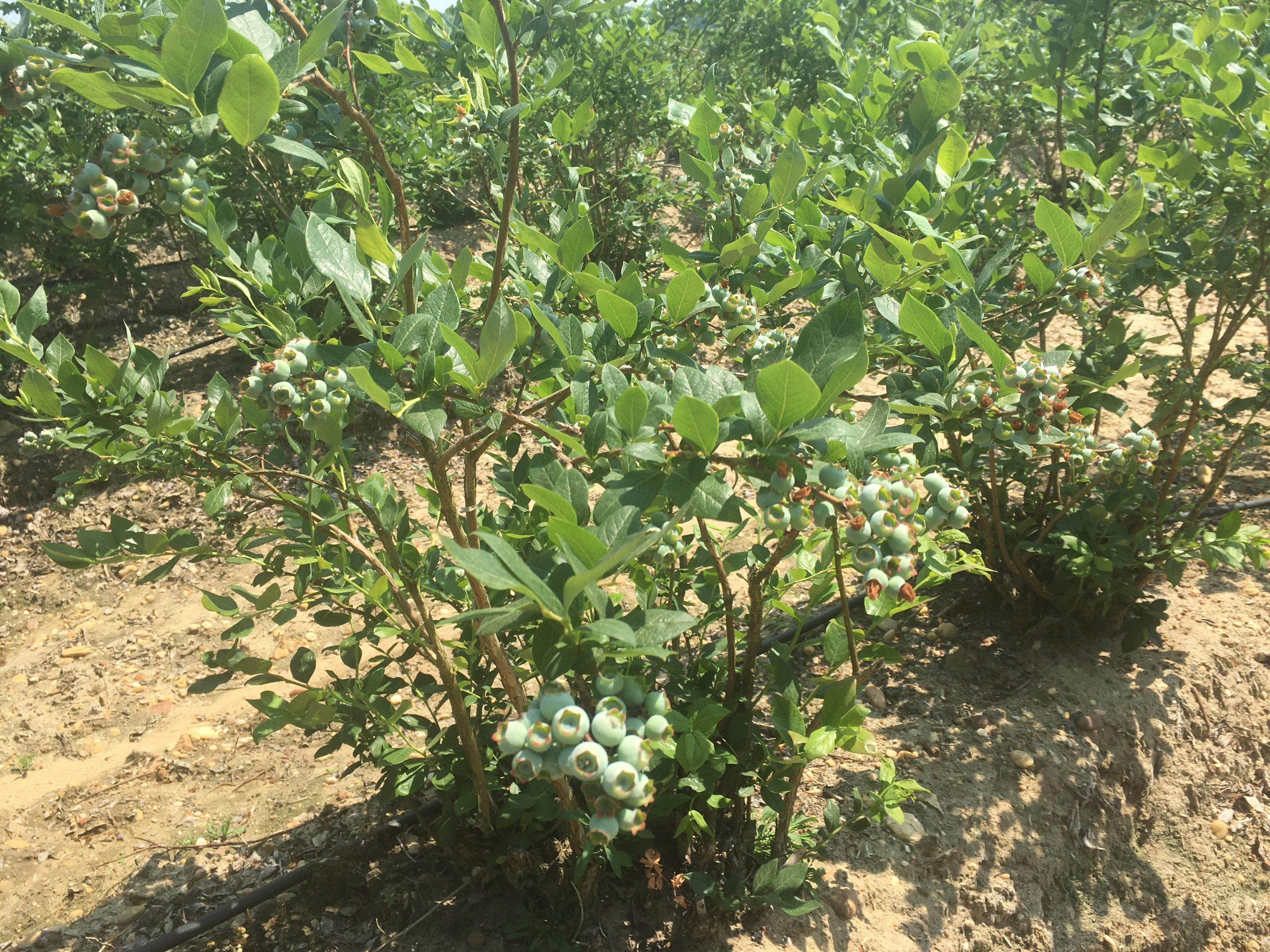}
\caption{iPhone}
\label{fig:handheld-iphone}
\end{subfigure}
\caption{Sample handheld images from Training Set.}
\end{figure}

A train/validation split of 84/16 was used to divide our data into a Training Set (80 images) and three Validation Sets A, B, C, each consisting of 5 images (see Table \ref{tab:berry-dataset}). We divided the Training Set into two parts (see Table \ref{tab:berry-training-dataset}): a Drone subset consists of 20 images captured by DJI Phantom 3 and Autel Evo II Pro drones, and a Handheld subset consisting of 60 images captured by various handheld cameras (Iphone, Android, Canon EOS Rebel).  All in the Training Set were taken in summer of 2021 and 2022 (see Figures \ref{fig:drone-dji}, \ref{fig:drone-evo}, \ref{fig:handheld-rebel}, \ref{fig:handheld-iphone} for sample images).  The validation datasets are different as follows: Validation Set A consists of 5 drone images taken in summer 2022 with a DJI Phantom 3 drone; Validation Set B consists of 5 drone images of the same foreground bushes as in Set A, but of their opposite side; Validation Set C consists of 5 drone images taken in Summer 2023 using a DJI Mini 3 drone.  

The reason for merging higher quality hand-held images with drone images to create our (Merged) Training Set when we began this project three years ago is that we anticipated future improvements in drone cameras to match the spatial resolution of current hand-held cameras (different cameras can have the same pixel resolution but different spatial resolution, even after accounting for ground sampling distance).  In particular, images captured with the DJI Mini 3 from 2023 (Validation Set C) appears to have higher spatial resolution in comparison to those captured with the DJI Phantom 3 in 2022 (Validation Set A).  We present validation metrics in Section VI that provides evidence to support this, although it should be noted that Validation Sets A and C are images of different blueberry varieties, Duke and Draper, respectively, and that the ratio of green to blue berries differs significantly for these two datasets, which we address in the next section.

% Berry Validation Dataset table
\setcounter{topnumber}{10}
\begin{table}[htbp]
\begin{center}
\begin{tabular}{|c|r|r|r|r|} 
\hline
BERRY & Images & Green & Blue  & Total \\
DATASETS & & Annotations & Annotations & Annotations \\
\hline Training Set & 80 & 61,680 & 17,471 & 79,151\\
\hline
Validation Set A & 5 & 11,328  & 1059  & 12,387 \\
\hline
Validation Set B & 5 & 8810  & 906  & 9716 \\ 
\hline
Validation Set C & 5 & 13,093 & 7060 & 20,153\\ 
\hline
TOTAL & 95 & 94,911 & 26,496 & 121,407 \\
\hline
\end{tabular}
\caption{Berry Datasets \label{tab:berry-dataset}}
\end{center}
\end{table}

% Berry Training Dataset table
\setcounter{topnumber}{10}
\begin{table}[htbp]
\begin{center}
\begin{tabular}{|c|c|r|r|r|} 
\hline
BERRY & Images & Green & Blue  & Total \\
TRAINING SET & & Annotations & Annotations & Annotations \\
\hline 
Drone & 20 & 31,372 & 4694 & 36,066 \\
\hline
Handheld & 60 & 30,308 & 12,777 & 43,085 \\
\hline
TOTAL (Merged) & 80 & 61,680 & 17,471 & 79,151\\
\hline
\end{tabular}
\caption{Berry Training Set \label{tab:berry-training-dataset}}
\end{center}
\end{table}

\subsection{Bush Datasets}
The Bush datasets consists of 256 drone images of blueberry bushes captured various different altitudes and camera angles (e.g., birds-eye vs side views; see Figures \ref{fig:bba} and \ref{fig:fba}.  We used a train/validation split of 90/10 to define our Training and Validation Sets (see Table \ref{tab:bush-dataset}).  

% Bush table
\begin{table}[htbp]
\begin{center}
\begin{tabular}{|c|r|r|} 
\hline
BUSH DATASET & Images & Bush Annotations \\
\hline 
Training Set & 473 & 2684 \\
\hline
Validation Set & 26 & 314 \\
\hline
TOTAL & 256 & 2998 \\
\hline
\end{tabular}
\caption{Bush Datasets \label{tab:bush-dataset}}
\end{center}
\end{table}

\subsection{Annotation}
% ---------------------
Images in our dataset were manually annotated using computer vision platforms Roboflow and CVAT, and involved a team of over 10 people (mostly undergraduate research students in our research lab).  

\vspace{5pt}

\begin{figure}[htp]
\centering
\begin{tabular}{c}
    \includegraphics[width=0.48\textwidth]{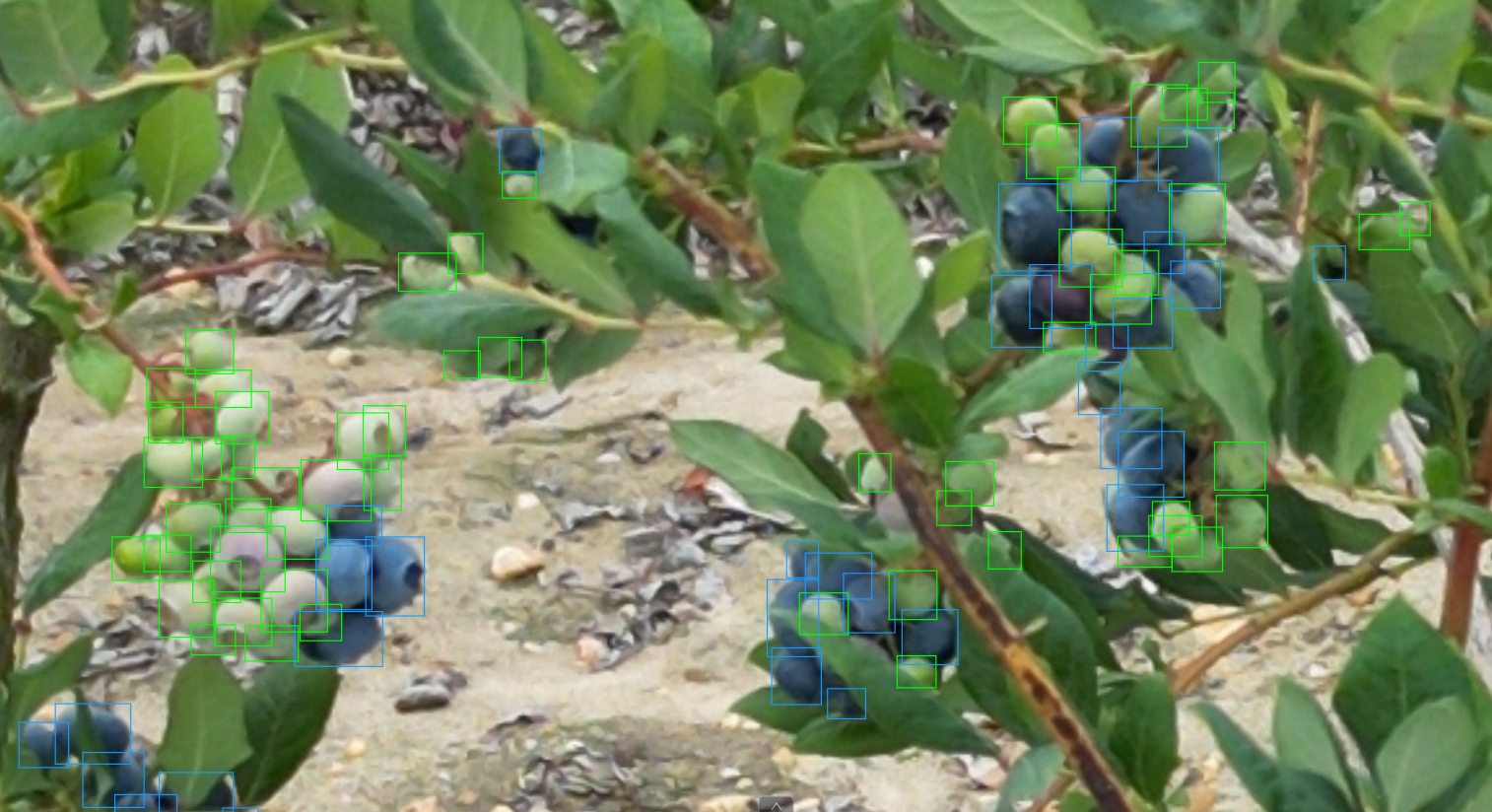}
\end{tabular}
\caption{Example of blueberry annotations (Green and Blue classes).}
\label{fig:ba}
\end{figure}

\begin{figure}[ht!]
\begin{subfigure}[b]{0.3\columnwidth}
\centering
\includegraphics[width=0.65\textwidth]{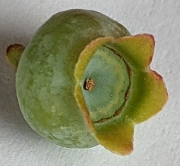}
%\caption{}
\label{fig:green-berry-1}
\end{subfigure}%
~
\begin{subfigure}[b]{0.3\columnwidth}
\centering
\includegraphics[width=0.65\columnwidth]{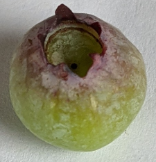}
%\caption{}
\label{fig:green-berry-2}
\end{subfigure}
~
\begin{subfigure}[b]{0.3\columnwidth}
\centering
\includegraphics[width=0.65\columnwidth]{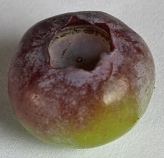}
%\caption{}
\label{fig:green-berry-3}
\end{subfigure}
\caption{Examples of berries labeled as Green}
\label{fig:green-berries}
\end{figure}

\begin{figure}[t!]
\begin{subfigure}[b]{0.3\columnwidth}
\centering
\includegraphics[width=0.65\textwidth]{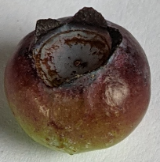}
%\caption{}
\label{fig:blue-berry-1}
\end{subfigure}%
~
\begin{subfigure}[b]{0.3\columnwidth}
\centering
\includegraphics[width=0.65\columnwidth]{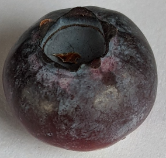}
%\caption{}
\label{fig:blue-berry-2}
\end{subfigure}
~
\begin{subfigure}[b]{0.3\columnwidth}
\centering
\includegraphics[width=0.65\columnwidth]{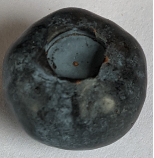}
%\caption{}
\label{fig:blue-berry-3}
\end{subfigure}
\caption{Examples of berries labeled as Blue}
\label{fig:blue-berries}
\end{figure}

\noindent \textbf{Berry Model}:
Images used for training our object detection Berry Model were annotated by drawing a rectangular bounding box tightly around each individual blueberry that is visible in the image and labeling it according to one of two color classes: Green or Blue (see Figure \ref{fig:ba}). 
We did not have an objective criteria for separating the berries into these two classes, except by providing the annotators with examples of berry images that were labeled by consensus (see Figures \ref{fig:green-berries} and \ref{fig:blue-berries}.  
We considered creating additional color classes to distinguish the berries, but decided on two classes to allow for quicker annotation given the total number of berry annotations in our dataset (over 120,000 annotations), which we believe to be the largest of its kind. The number of annotations (Green, Blue, total) are given in Table \ref{tab:berry-dataset}.  Observe that the number of Green annotations is significantly higher than the number of Blue annotations, with the ratio of Green to Blue highest for Validation Set A (10.7) and smallest for Validation Set C (1.85).  This is due to the fact that images in Set C were captured much closer to the first harvest date compared to the images in Set A.

\begin{figure}[t!]
\begin{subfigure}[b]{0.48\columnwidth}
\centering
\includegraphics[width=0.45\textwidth]{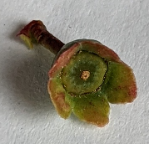}
%\caption{}
\label{fig:berry-too-small-1}
\end{subfigure}%
~
\begin{subfigure}[b]{0.5\columnwidth}
\centering
\includegraphics[width=0.35\columnwidth]{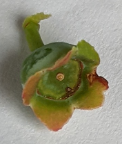}
%\caption{}
\label{fig:berry-too-small-2}
\end{subfigure}
\caption{Examples of berries considered too small (not annotated)}
\label{fig:berries-too-small}
\end{figure}

 Occluded berries were annotated if the annotator was convinced that the object was a berry and only its visible portion was annotated. Shadowy and/or blurry berries were also annotated according to the same criteria.  We acknowledge that this criteria is dependent on the visual acuity of the annotator and
exposes the difficulty of annotating tiny objects such as blueberries. One can of course utilize majority voting by employing a group of annotators; however, given the large number of berries in a single image (over 1000 berries were annotated on average), we considered this approach to not be feasible given our limited resources.
In addition, we annotated only those berries that are large enough to be \textit{harvestable}. Berries that fruited late in the season are too small in size to be harvestable by commercial farms. Figure \ref{fig:berries-too-small}) shows examples of berries that were not annotated.

\vspace{5pt}

\begin{figure}[htp]
\centering
\begin{tabular}{c}

\includegraphics[width=0.48\textwidth]{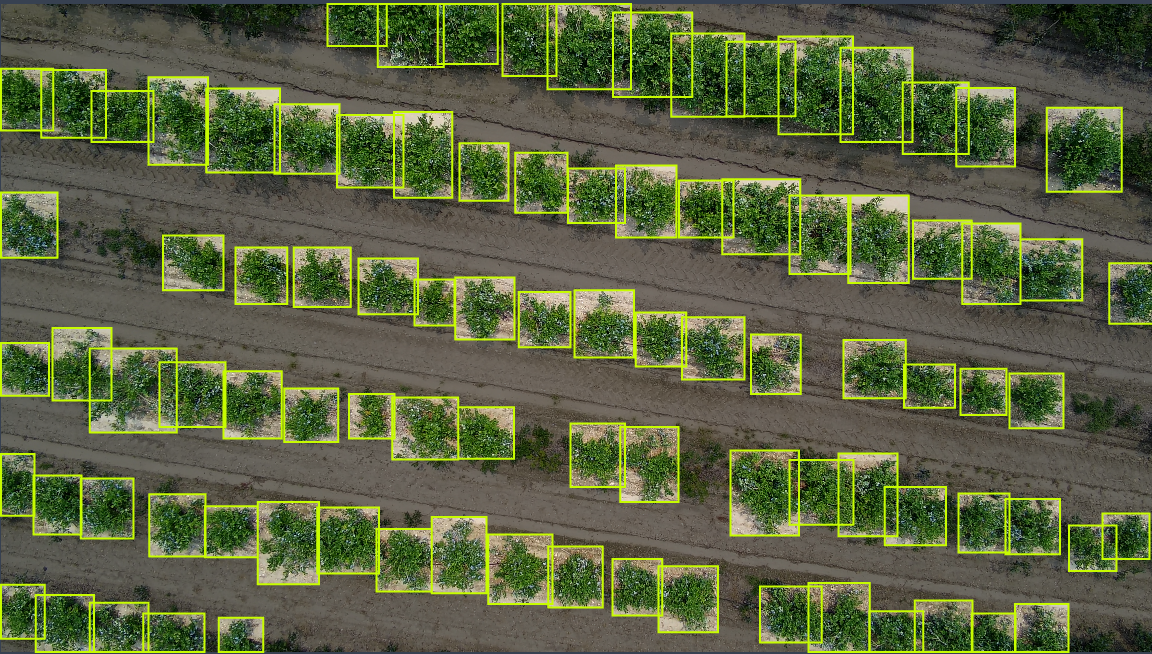}
\end{tabular}
\caption{Example of bush annotations (birds-eye view).}
\label{fig:bba}
\end{figure}

\begin{figure}[htp]
\centering
\begin{tabular}{c}
    \includegraphics[width=0.48\textwidth]{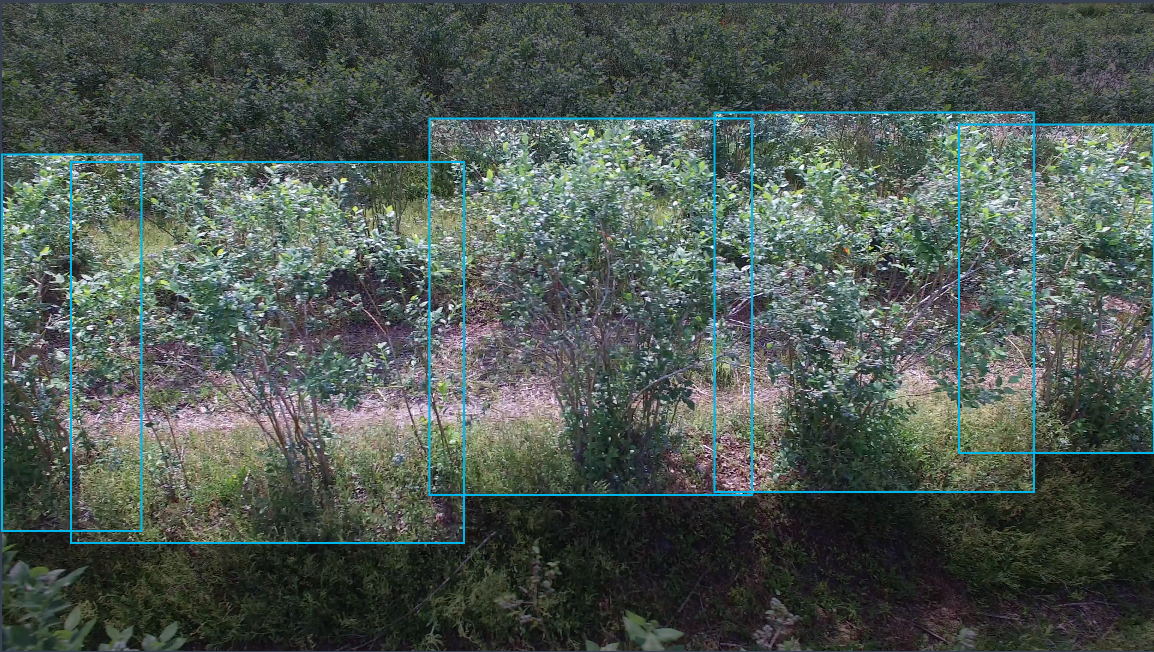}
\end{tabular}
\caption{Example of foreground bush annotations (side view).}
\label{fig:fba}
\end{figure}

\noindent \textbf{Bush Model}:
For our Bush Detection model, images were annotated by drawing a rectangular bounding box around each blueberry bush in the foreground of the image, including its trunk (if visible) and all branches (see Figures \ref{fig:bba} and \ref{fig:fba}).  Neighboring bushes with long branches will unavoidably cause their bounding boxes to overlap as seen in Figure \ref{fig:fba}.

\subsection{Data Augmentation}
% ---------------------
We perform data augmentation as follows to create diverse training dataset for more effective training. For our bush model, we applied variations in Hue (+/- 0.015),
Saturation (+/- 0.7),
Value (brightness) (+/- 0.4)
Translation (+/- 0.1),
Scale (+/- 0.5), and 
Flip Left-Right (0.5 probability).

\begin{figure}[htp]
\centering
\begin{tabular}{c}
    \includegraphics[width=0.48\textwidth]{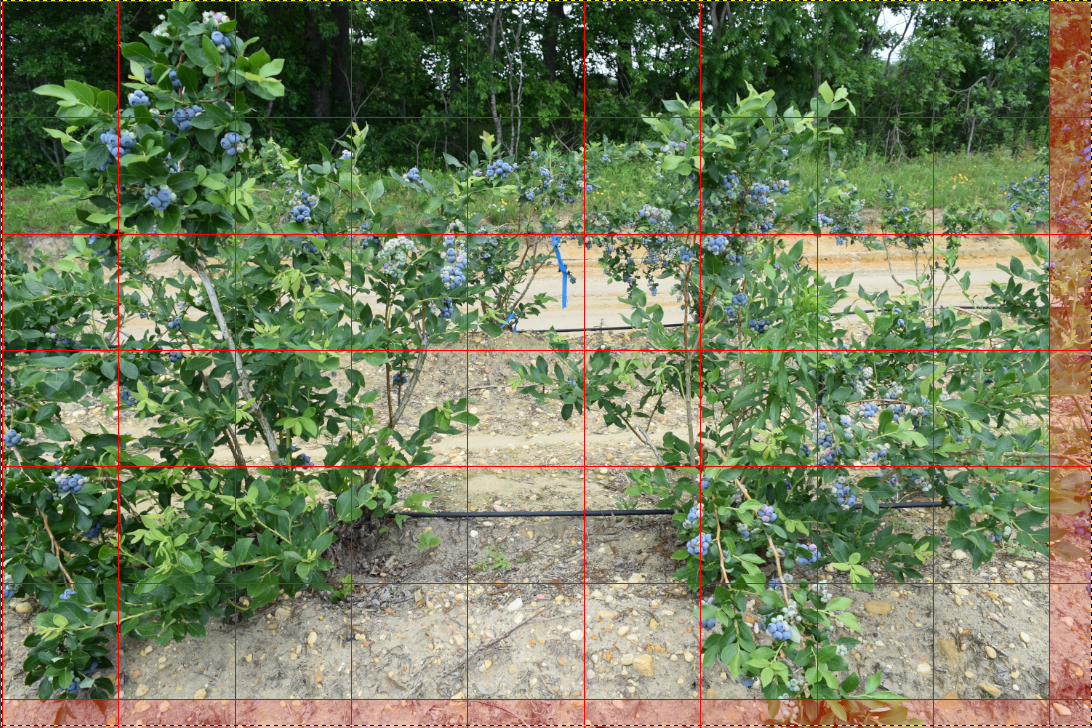}
\end{tabular}
\caption{Depiction of the tiling process.}
\label{fig:tiling1}
\end{figure}

However, for our berry counting model, data augmentation for its the training dataset consisted of dividing each original full-size image into tiles. 
This is because the YOLO model works better with smaller, square images; thus, we divide each image into $640\times 640$ tiles. If the image cannot be perfectly cut into $640\times 640$ (starting at top left), then the leftover pieces are discarded. See Figure \ref{fig:tiling1} for a depiction of the tiling process.  Any berry annotation with its bounding box appearing in two neighboring tiles is decided by choosing the tile that contains the center of the box.

\section{YOLO Objection Detection}
% -----------------------------------------------

\subsection{YOLOv5}
% ---------------------
YOLOv5 is the fifth version of the YOLO (You Only Look Once) family of compound-scaled object detection models, first introduced by Redmon et al in 2015 \cite{Redmon_2016_CVPR}. We chose YOLOv5 over other object-detection models such as Faster R-CNN because of its state-of-the-art performance (back when we began this project over three years ago) in terms of inference speed relative to accuracy.  There are of course many higher versions of YOLO now available; however, experiments show negligible improvement in training metrics for both the Berry and Bush models when using say YOLOv8 in comparison to YOLOv5.  The focus of this paper is not to present a comparison of all the different versions of YOLO, but to establish baseline results for YOLOv5 that others can compare their results against. 

Both Berry and Bush Models were trained using YOLOv5s (small) since there were little improvement in training metrics when resorting to larger-sized models such as YOLOv5m (medium) and YOLOv5l (large).  In addition, especially for the Bush Model, where inference needs to be performed in real time by the drone on-board computer during flight, the small size model makes this possible.  We were able to achieve 6-9 frames per second performing detection the Bush model on a Jetson Nano with Deepstream running YOLOv5s.

\subsection{Training}

Both the Bush and Berry Models were trained on Bush and Berry Datasets, respectively, using  YOLOv5s' default hyperparameter settings found in its hyperparameter yaml file "hyp.scratch-low.yaml". We used a batch size of 32 and trained for up to 400 epochs for the Bush Model and 300 epochs for the Berry Model; the default early stopping criterion was used to stop training when cross-validation loss diverged from the training loss.  Training metrics were calculated using Ultralytics YOLOv5 utils Python library through the script 'metric.py'. We found little difference in accuracy by changing other default parameters and believe optimizing these parameters would not change the conclusions of our paper; we believe that it is more important to increase the size of our dataset and add more higher-resolution images in order to significantly improve our model.

\subsection{Validation}

The best fold from five-fold cross-validation of each model (base on highest mAP:0.5) was used to perform validation on the datasets in Tables \ref{tab:berry-dataset} and \ref{tab:bush-dataset}.  For the Berry Model, each image was first divided into overlapping $700\times 700$ tiles, or as close as possible to these dimensions, so that they overlap by 60 pixels in each dimension; this avoids double-counting berries that may be split if non-overlapping tiles were used.  Each tile was then passed through the Berry Model and post-processing was used to remove duplicate bounding boxes that appear in two overlapping tiles.  To compute precision and recall for each class, a confidence threshold of 0.1 was applied to generate detections and an IOU threshold of 0.3 was applied to count true positives when comparing them against ground truths; if more than one detection matched a ground truth in terms of both IOU threshold and class, then the detection with the highest confidence is selected.  A low IOU threshold was used (in comparison to YOLOv5's default threshold of 0.6) to avoid eliminating correct predictions that did not overlap sufficiently with the ground truth.  This is because bounding boxes of blueberries are small in dimension; thus, detection errors in the position of these boxes by just a few pixels can significant impact their IOU.

\subsection{DeepSORT Tracking}

To test the accuracy of our Bush Model in tracking bushes as discussed in our pipeline in Section 3, we used DeepSORT to calculate multiple object tracking accuracy (MOTA).  DeepSORT is a computer vision tracking algorithm for tracking objects from a video stream by assigning an ID to each detected \cite{deepsort}.  It is an extension of the Simple Online and Realtime Tracking (SORT) because it  integrates appearance information based on a deep appearance descriptor.  We applied DeepSORT to two short video clips: one of a drone performing Point Sampling and the other performing Row Sampling.  Results are presented in Section VI.

\section{Experimental Results}
% -----------------------------------------------

In this section we present training and validation results  for our Berry and Bush Models, both separately and also when combined to perform bush-cropping in order to obtain a total berry count for only the foreground center bush.  We also present tracking results (MOTA) for the Bush Model using the DeepSORT algorithm.

\subsection{Berry Model}
% ---------------------

 Three different Berry Models were trained using five-fold cross-validation: Drone, Handheld, and Merged.  The Drone and Handheld Berry Models were trained on the 20 drone images and 60 handheld images, respectively (see Table \ref{tab:berry-training-dataset}).  The Merged Berry Model (or simply Berry Model) was trained on the merged dataset of 80 images (drone and handheld); we refer to this merged dataset as the Berry Training Set.

\vspace{5pt}
\noindent \textbf{Training Results}: Training metrics for the three Berry Models (Drone, Handheld, Merged) are given in Tables \ref{tab:training-metrics-drone}-\ref{tab:training-metrics-merged}, respectively.  The Handheld Model performed best across all metrics as expected (precision, recall, mAP, and F1) since the bushes in handheld images were shot at a closer distance with berries appearing larger than those in drone images, thus making detection easier. Drone Model performed almost as well as the Handheld Model in terms of precision, but recall was significantly worse.  Of course, these two models are only as good as the data that they are trained on, and so when we present validation results below we'll see their performance is reversed, thus supporting the need for a Merged Model that is robust to a variety of different types of images.

% Drone Training Table
%\FloatBarrier
\setcounter{topnumber}{10}
\begin{table}[htbp!]
\begin{center}
\begin{tabular}{|c|c|c|c|c|c|} 
\hline
Training & Precision & Recall & mAP 0.5  & mAP 0.5:0.95 & F1\\
\hline 
%\verb|Blueberry_Berry_Drone_20_6-8-2024_640x640_Fold1| 
Fold 1&0.8271&0.6472&0.7281&0.3773&0.7262\\
\hline 
%\verb|Blueberry_Berry_Drone_20_6-8-2024_640x640_Fold2| 
Fold 2&0.8039&0.6378&0.7143&0.3724&0.7113\\
\hline 
%\verb|Blueberry_Berry_Drone_20_6-8-2024_640x640_Fold3| 
Fold 3&0.8184&0.6692&0.7350&0.3848&0.7363\\
\hline 
%\verb|Blueberry_Berry_Drone_20_6-8-2024_640x640_Fold4|
Fold 4&0.8347&0.6583&0.7446&0.3945&0.7361\\
\hline 
%\verb|Blueberry_Berry_Drone_20_6-8-2024_640x640_Fold5|
Fold 5&0.8032&0.6707&0.7295&0.3753&0.7310\\
\hline
\hline
%\verb|MEAN:| 
Mean&0.8175&0.6566&0.7303&0.3809&0.7282\\
\hline
%\verb|STANDARD DEVIATION:| 
SD & 0.0139&0.0142&0.0110&0.0089&0.0103\\
\hline
\end{tabular}
\caption{Training metrics for the Drone Berry Model \label{tab:training-metrics-drone}}
\end{center}
\end{table}
%\FloatBarrier

% Handheld Training Table
\setcounter{topnumber}{10}
\begin{table}[htbp!]
\begin{center}
\begin{tabular}{|c|c|c|c|c|c|} 
\hline
Training & Precision & Recall & mAP 0.5  & mAP 0.5:0.95 & F1\\
\hline 
%\verb|Blueberry_Berry_Handheld_60_6-8-2024_640x640_Fold1| 
Fold 1&0.8230&0.7518&0.8159&0.5119&0.7858\\
\hline 
%\verb|Blueberry_Berry_Handheld_60_6-8-2024_640x640_Fold2| 
Fold 2&0.8527&0.7623&0.8339&0.5290&0.8050\\
\hline 
%\verb|Blueberry_Berry_Handheld_60_6-8-2024_640x640_Fold3| 
Fold 3&0.8553&0.7450&0.8287&0.5176&0.7963\\
\hline 
%\verb|Blueberry_Berry_Handheld_60_6-8-2024_640x640_Fold4|
Fold 4&0.8606&0.8009&0.8639&0.5418&0.8297\\
\hline 
%\verb|Blueberry_Berry_Handheld_60_6-8-2024_640x640_Fold5|
Fold 5&0.8259&0.7640&0.8269&0.5320&0.7938\\
\hline
\hline
%\verb|MEAN:| 
Mean&0.8435&0.7648&0.8338&0.5264&0.8021\\
\hline
%\verb|STANDARD DEVIATION:|
SD & 0.0177&0.0216&0.0180&0.0119&0.0169\\
\hline
\end{tabular}
\caption{Training metrics for the Handheld Berry Model \label{tab:training-metrics-handheld}}
\end{center}
\end{table}

% Merged Training Table
\setcounter{topnumber}{10}
\begin{table}[htbp!]
\begin{center}
\begin{tabular}{|c|c|c|c|c|c|} 
\hline
Training & Precision & Recall & mAP 0.5  & mAP 0.5:0.95 & F1\\
\hline 
%\verb|Blueberry_Berry_Merged_80_6-11-2024_640x640_Fold1|
Fold 1&0.8434&0.7267&0.7965&0.4822&0.7807\\
\hline 
%\verb|Blueberry_Berry_Merged_80_6-11-2024_640x640_Fold2|
Fold 2&0.8303&0.7160&0.7852&0.4643&0.7690\\
\hline 
%\verb|Blueberry_Berry_Merged_80_6-11-2024_640x640_Fold3|
Fold 3&0.8477&0.7253&0.7941&0.4805&0.7817\\
\hline 
%\verb|Blueberry_Berry_Merged_80_6-11-2024_640x640_Fold4|
Fold 4&0.8323&0.7260&0.7962&0.4806&0.7755\\
\hline 
%\verb|Blueberry_Berry_Merged_80_6-11-2024_640x640_Fold5|
Fold 5&0.8319&0.7085&0.7820&0.4636&0.7653\\
\hline
\hline
%\verb|MEAN:|
Mean&0.8371&0.7205&0.7908&0.4742&0.7744\\
\hline
%\verb|STANDARD DEVIATION:|
SD & 0.0079&0.0080&0.0067&0.0094&0.0072\\
\hline
\end{tabular}
\caption{Training metrics for the Merged Berry Model  \label{tab:training-metrics-merged}}
\end{center}
\end{table}

\noindent \textbf{Validation Results}:
Tables \ref{tab:re-val-metrics-three-berry-models-fullsize-set-a}, \ref{tab:re-val-metrics-three-berry-models-fullsize-set-b}, and \ref{tab:re-val-metrics-three-berry-models-fullsize-set-c} show precision and recall for the three Berry Models (Drone, Handheld, Merged) validated on Validation Sets A, B, and C, respectively.  For Sets A and B (Tables \ref{tab:re-val-metrics-three-berry-models-fullsize-set-a}, \ref{tab:re-val-metrics-three-berry-models-fullsize-set-b}), the Drone Berry Model had the highest overall precision among all models; however, the Merged Model had significantly higher overall recall and slightly better Blue precision. On the other hand, the Handheld Berry Model performed the worst in all categories.  In particular, Green recall was extremely low, indicating that the model failed to detect many green berries, especially those in background bushes where they appear much smaller, which makes it more difficult for the model to difficult them.  Also, precision for class Blue was also quite low, which shows the Handheld Berry Model did a poor job of correctly detecting blue berries. 

% Re-Annotated Validation Metrics for Full, Drone, Handheld Berry Models - Full-Size Images, Set A
%\FloatBarrier
\setcounter{topnumber}{10}
\begin{table}[htbp!]
\begin{center}
\begin{tabular}{|c|c|c|c|c|c|c|} 
\hline
Berry & Prec. & Rec. &  Prec. & Rec. & Prec. & Rec. \\
Model & (Green) & (Green) & (Blue) & (Blue) & (Overall)   & (Overall)   \\
\hline 
Drone & 0.775 &	0.708 &	0.807 &	0.245 &	0.776 &	0.67   \\
\hline 
Handheld & 0.749 &	0.069 &	0.092 &	0.283 &	0.255 &	0.087  \\
\hline
Merged & 0.755 &	0.745 &	0.804 &	0.361 &	0.757 &	0.713  \\
\hline
\end{tabular}
\caption{Drone, Handheld, Merged Berry Models: Validation metrics for Validation Set A
\label{tab:re-val-metrics-three-berry-models-fullsize-set-a}}
\end{center}
\end{table}

% Re-Annotated Validation Metrics for Full, Drone, Handheld Berry Models - Full-Size Images, Set B
%\FloatBarrier
\setcounter{topnumber}{10}
\begin{table}[htbp!]
\begin{center}
\begin{tabular}{|c|c|c|c|c|c|c|} 
\hline
Berry & Prec. & Rec. &  Prec. & Rec. & Prec. & Rec. \\
Model & (Green) & (Green) & (Blue) & (Blue) & (Overall)   & (Overall)   \\
\hline 
Drone & 0.657 &	0.675 &	0.621 &	0.245 &	0.655 &	0.637   \\
\hline 
Handheld & 0.608 &	0.205 &	0.111 &	0.461 &	0.337 &	0.228  \\
\hline
Merged &  0.601	& 0.734 &	0.69 &	0.317 &	0.605 & 0.697 \\
\hline
\end{tabular}
\caption{Drone, Handheld, Merged Berry Models: Validation metrics for Validation Set B
\label{tab:re-val-metrics-three-berry-models-fullsize-set-b}}
\end{center}
\end{table}

As for Validation Set C (Table \ref{tab:re-val-metrics-three-berry-models-fullsize-set-c}), performance reversed with the Merged Model now having the highest overall precision and only slightly worse overall recall compared to the Drone and Handheld Models.  Observe that overall recall the Handheld Model significantly improved compared to results in Tables \ref{tab:re-val-metrics-three-berry-models-fullsize-set-a}, \ref{tab:re-val-metrics-three-berry-models-fullsize-set-b}.  However, all three models performed poorly on overall recall.  An inspection of the false negatives were of berries on background bushes that were annotated but either too small for any of the models to detect or too shaded for the model to distinguish as a berry.  The results on overall precision provide evidence that training on combined drone and handheld images helped to improve the (Merged) Berry Model, with only a slight decrease in overall recall (but best Green recall), in comparison to training on drone and handheld images separately.

% Re-Annotated Validation Metrics for Full, Drone, Handheld Berry Models - Full-Size Images, Set C
%\FloatBarrier
\setcounter{topnumber}{10}
\begin{table}[htbp!]
\begin{center}
\begin{tabular}{|c|c|c|c|c|c|c|} 
\hline
Model & (Green) & (Green) & (Blue) & (Blue) & (Overall)   & (Overall)   \\
Model & (Green) & (Green) & (Blue) & (Blue) &   &   \\
\hline 
Drone &  0.609 &	0.478 &	0.866 &	0.433 &	0.68 &	0.461  \\
\hline 
Handheld & 0.793 &	0.459 &	0.818 &	0.476 &	0.802 &	0.465  \\
\hline
Merged & 0.815 &	0.462 &	0.872 &	0.434 &	0.835 &	0.451  \\
\hline
\end{tabular}
\caption{Drone, Handheld, Merged Berry Models: Validation metrics for Validation Set C
\label{tab:re-val-metrics-three-berry-models-fullsize-set-c}}
\end{center}
\end{table}

Table \ref{tab:re-val-metrics-merged-berry-model-fullsize-sets-a-b-c} isolates precision and recall for only the Merged Berry Model to help better compare results for the three validation sets (A, B, and C).  Results for Set C yielded the highest overall precision, but unfortunately also yielded the lowest overall recall, which we previously described as due to berries on background bushes that are too small in terms of pixel resolution for the model to detect.  This is supported by results that we will present later on when we consider detecting berries only the foreground center bush.

\begin{figure}[htbp!]
\centering
\begin{tabular}{c}
    \includegraphics[width=0.48\textwidth]{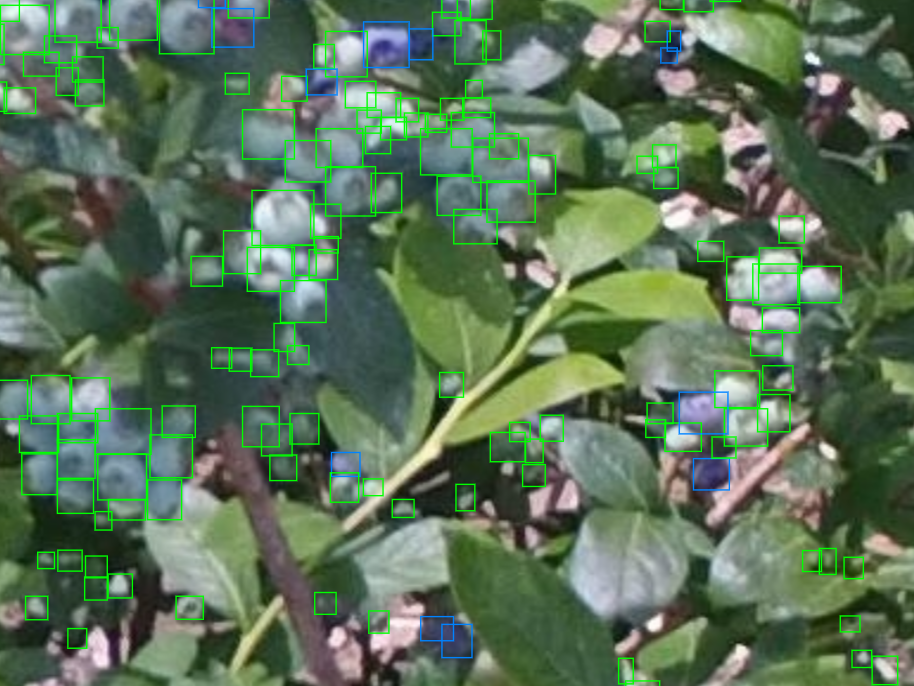}
\end{tabular}
\caption{Close-up view of ground truth annotations in sample image from Validation Set B (green and blue boxes denote Green and Blue classes, respectively).}
\label{fig:val-gt}
\end{figure}

\begin{figure}[htbp!]
\centering
\begin{tabular}{c}
    \includegraphics[width=0.48\textwidth]{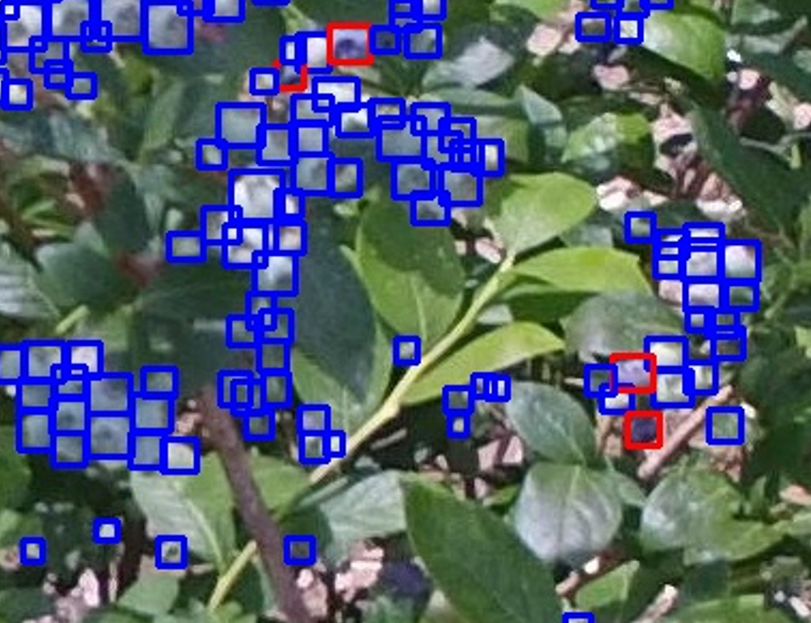}
\end{tabular}
\caption{Close-up view of Berry Model predictions in sample image from Validation Set B (blue and red boxes denote Green and Blue classes, respectively).}
\label{fig:val-pred}
\end{figure}

\begin{figure}[htbp!]
\centering
\begin{tabular}{c}
    \includegraphics[width=0.48\textwidth]{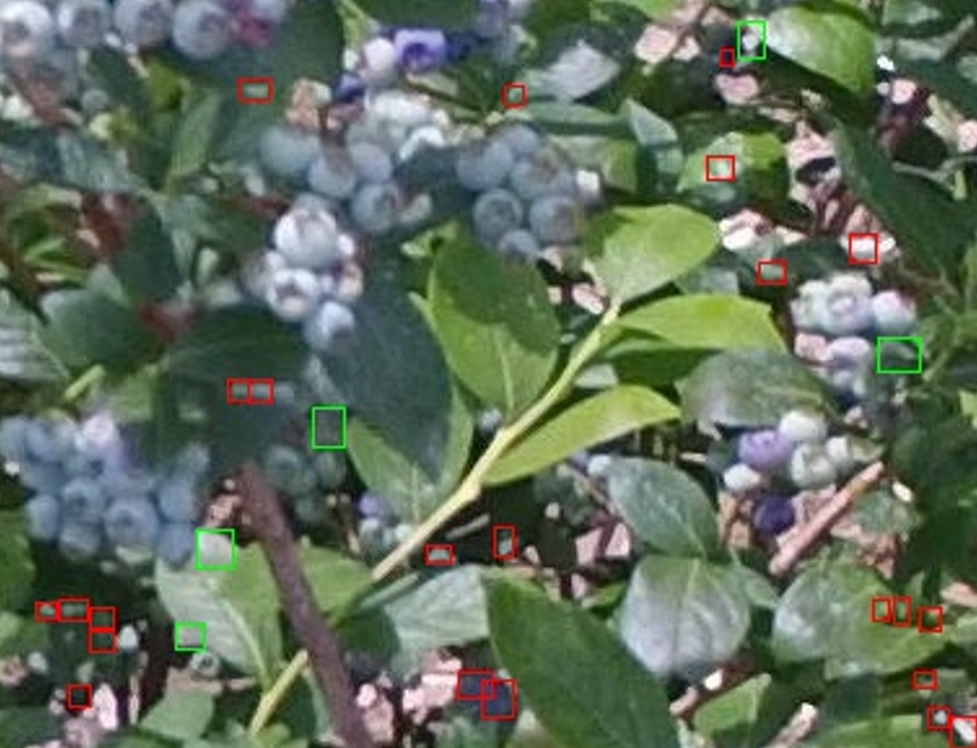}
\end{tabular}
\caption{Close-up view of Berry Model false positives (green) and false negatives (red) for sample image in Validation Set B.}
\label{fig:val-fp-fn}
\end{figure}

Conversely, results for Set B had the lowest overall precision, but highest overall recall. 
Figures \ref{fig:val-gt}, \ref{fig:val-pred}, and \ref{fig:val-fp-fn} show a sample image (B2) from Set B, but with different types of bounding boxes drawn: ground truths (Figure \ref{fig:val-gt}), predictions (Figure \ref{fig:val-pred}), and false positives and false negatives (Figure \ref{fig:val-fp-fn}).  A close inspection of the false positives shows that some of them could possibly be berries, but difficult to discern clearly, which explains why they were not annotated. 

% Re-Annotated Validation Metrics for Full Berry Model - Full-Size Images, Sets A,B,C
%\FloatBarrier
\setcounter{topnumber}{10}
\begin{table}[htbp!]
\begin{center}
\begin{tabular}{|c|c|c|c|c|c|c|} 
\hline
Validation & Prec. & Rec. &  Prec. & Rec. & Prec. & Rec. \\
Dataset & (Green) & (Green) & (Blue) & (Blue) & (Overall)   & (Overall)  \\
\hline 
Set A & 0.755 &	0.745 &	0.804 &	0.361 &	0.757 &	0.713   \\
\hline 
Set B & 0.601 &	0.734 &	0.69 &	0.317 &	0.605 &	0.697  \\
\hline
Set C & 0.815 &	0.462 &	0.872 &	0.434 &	0.835 &	0.451   \\
\hline
\end{tabular}
\caption{Merged Berry Model validation metrics for Validation Sets
A, B, C
\label{tab:re-val-metrics-merged-berry-model-fullsize-sets-a-b-c}}
\end{center}
\end{table}

\subsection{Bush Model}
% ---------------------
\noindent \textbf{Training/Validation Results}: Table \ref{tab:training-metrics-bush} gives training metrics for the Bush Model trained on the Bush Training Set.  Table \ref{tab:validation-results-bush} gives validation metrics for the Bush Model validated on the Bush Validation Set.  Both tables show high precision at around 90\%, and good recall ranging from high 70\% for training to low 80\% for validation.  A review the true negatives (bushes that were not detected) indicates that the Bush Model struggled to detect those bushes at the edge of the image.  Fortunately, this issue is not a concern since the goal of the Bush Model is to detect and track foreground bushes.

\begin{figure}[htbp!]
\centering
\begin{tabular}{c}
    \includegraphics[width=0.48\textwidth]{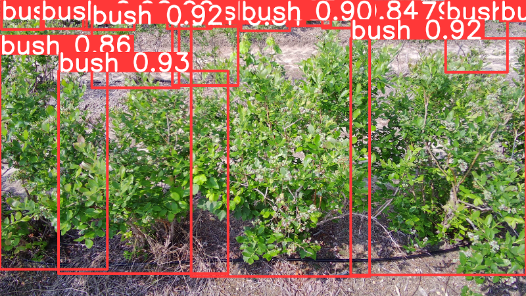}
\end{tabular}
\caption{Bush Model prediction on a sample validation image (angled-side view of bush).}
\label{fig:val-pred_1}
\end{figure}

\begin{figure}[htbp!]
\centering
\begin{tabular}{c}
    \includegraphics[width=0.48\textwidth]{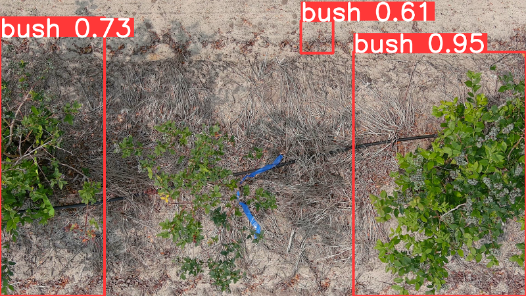}
\end{tabular}
\caption{Bush Model prediction on a sample validation image (birds-eye view) showing a false positive (bush marked with confident 0.61) and false negative (bush tagged with blue ribbon).}
\label{fig:val-pred_2}
\end{figure}

Figures \ref{fig:val-pred_1} and \ref{fig:val-pred_2} show sample predictions of the Bush Model from an angled-side view and birds-eye view, respectively, including a false positive and false negative in the latter figure. 

% Bush Detection Low-Altitude Angle-View Training Results Table
\setcounter{topnumber}{10}
\begin{table}[htbp!]
\begin{center}
\begin{tabular}{|c|c|c|c|c|} 
\hline
Training Fold & Precision & Recall & mAP 0.5  & mAP 0.5:0.95 \\
\hline 
Fold 1 & 0.899	&	0.76	&	0.867	&	0.508\\
\hline 
Fold 2 & 0.842	&	0.723	&	0.807	&	0.433\\
\hline 
Fold 3 & 0.881	&	0.774	&	0.869	&	0.493\\
\hline 
Fold 4 & 0.939	&	0.893	&	0.946	&	0.592\\
\hline 
Fold 5 & 0.877	&	0.733	&	0.829	&	0.492\\
\hline
\hline
%\verb|MEAN:| 
Mean & 0.888	&	0.777	&	0.864	&	0.504\\
\hline
%\verb|STANDARD DEVIATION:| 
SD & 0.035 &	0.068	&	0.053	&	0.057\\
\hline
\end{tabular}
\caption{Training metrics for Bush Model trained on Bush Validation Set}
\label{tab:training-metrics-bush}
\end{center}
\end{table}

% Bush Detection Low-Altitude Angle-View Validation Results Table
\setcounter{topnumber}{10}
\begin{table}[htbp!]
\begin{center}
\begin{tabular}{|c|c|c|c|c|} 
\hline
Dataset & Precision & Recall & mAP 0.5  & mAP 0.5:0.95 \\
\hline 
Bush Validation Set & 0.916	&	0.834	&	0.916	&	0.538\\
\hline 
\end{tabular}
\caption{Validation metrics for Bush Model validated on Bush Validation Set (using best fold) \label{tab:validation-results-bush}}
\end{center}
\end{table}

\begin{figure}[htbp!]
\centering
\begin{tabular}{c}
    \includegraphics[width=0.48\textwidth]{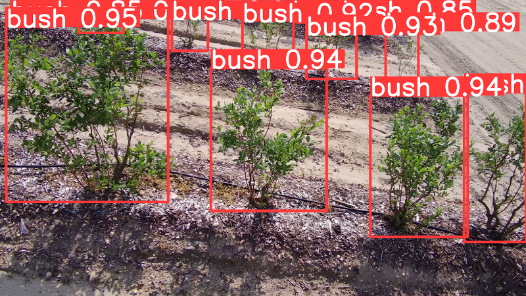}
\end{tabular}
\caption{Bush Model prediction on an example validation image (slanted view).}
\label{fig:val-pred_3}
\end{figure}

\vspace{5pt}

\noindent \textbf{Bush Tracking}: We calculated Multiple Object Tracking Accuracy (MOTA) for two video clips: Bush Video 1 and Bush Video 2.  Bush Video 1 (24 seconds) was captured by a DJI drone flying overhead to a bush and simultaneously adjusting its position and camera angle from birds-eye to angled-side view. Bush Video 2 (5 seconds) was captured by a DJI drone flying sideways along a row of blueberry bushes.  Table \ref{tab:bush-video-mota} gives MOTA results for both video clips.  Although MOTA is lower that what we hoped for, a review of the detections shows that the Bush Model does a very good of tracking the foreground center bush, which is the primary goal of the model, and performs worse for bushes at the edges of the image, something that we previously mentioned.

% Bush MOTA

\setcounter{topnumber}{10}
\begin{table}[htbp!]
\begin{center}
\begin{tabular}{|c|c|c|} 
\hline
PARAMETER & BUSH VIDEO 1 & BUSH VIDEO 2 \\
\hline
Number of Frames & 365 & 75 \\
\hline
Bush Annotations & 6464 & 324 \\
\hline
Predictions & 7104 & 352 \\
\hline
Mismatch Errors & 149 & 7 \\
\hline
False Positives & 2315 & 96 \\
\hline
False Negatives & 1745 & 66 \\
\hline
IOU Threshold & 0.5 & 0.5 \\
\hline
\textbf{MOTA} & \textbf{0.3489} & \textbf{0.4786} \\
\hline 
\end{tabular}
\caption{MOTA for Bush Model of Videos 1 and 2 \label{tab:bush-video-mota}}
\end{center}
\end{table}

\subsection{Bush-Cropped Berry Model}
\begin{figure}[htbp!]
\begin{subfigure}[b]{0.48\columnwidth}
\centering
\includegraphics[width=1\textwidth]{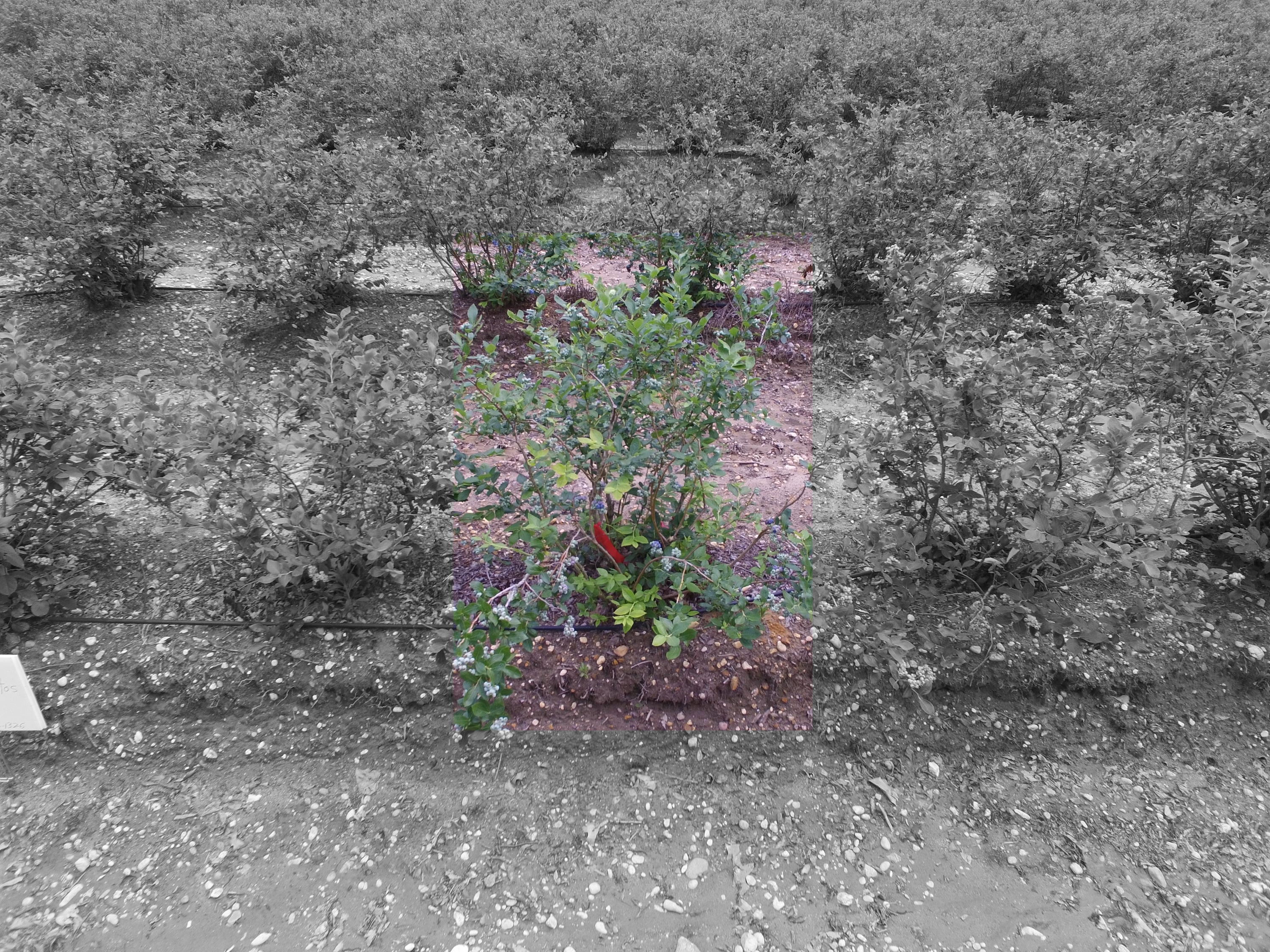}
\caption{Full image}
\label{fig:full-bush}
\end{subfigure}%
~
\begin{subfigure}[b]{0.5\columnwidth}
\centering
\includegraphics[width=0.55\columnwidth]{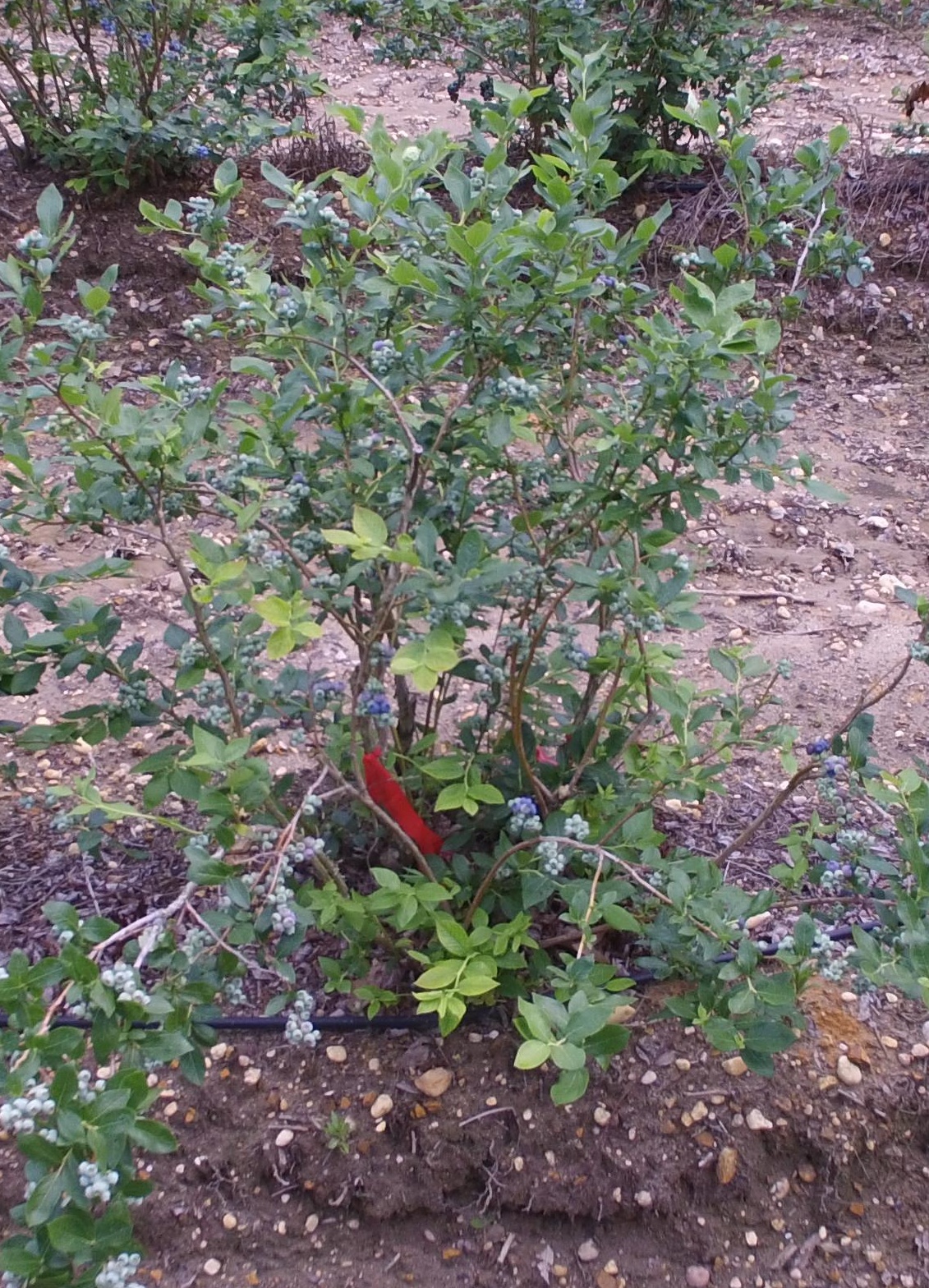}
\caption{Cropped image}
\label{fig:cropped-bush}
\end{subfigure}
\caption{Example of image cropped around foreground center bush by Bush Model and then fed into Berry Model.}
\end{figure}

%\subsection{Set-up}
% ---------------------
Together, the Berry and Bush Models can be combined into a pipeline to detect only those berries that appear in a single bush.  We call this pipeline the Bush-Cropped Berry Model.  In particular, we first pass a full-size image through the Bush Model to obtain an array of detected bushes and their corresponding bounding boxes. From these bounding boxes we select one called the \textit{central bounding box} (corresponding to the foreground center bush; see Figures \ref{fig:full-bush} and \ref{fig:cropped-bush}), whose center is closest (in terms of radial distance) to the center of the image, and then crop the image (using OpenCV2) around the central bounding box.  The cropped image is then passed through the Berry Model to detect berries.  Detections are compared against those ground truths contained within the central bounding box. 

Tables \ref{tab:re-val-metrics-cropped-set-a}, \ref{tab:re-val-metrics-cropped-set-b}, and \ref{tab:re-val-metrics-cropped-set-c} show validation results using the Bush-Cropped Berry Model (Drone, Handheld, and Merged) on Validation Sets A, B, and C, respectively.  
This time we see the Merged Model outperforming the Drone and Handheld Models in overall recall for all validation sets.  The Handheld Berry Model performed the worst in both overall precision and recall for Sets A and B (Tables \ref{tab:re-val-metrics-cropped-set-a} and \ref{tab:re-val-metrics-cropped-set-b}), as already observed when validated on entire images (see Table \ref{tab:re-val-metrics-merged-berry-model-fullsize-sets-a-b-c}), but performed surprising well on Green precision for Set B (Table \ref{tab:re-val-metrics-cropped-set-b}).  

For Set C, the Merged Model achieved the best overall precision and recall, with the Handheld Model having only slightly worse results.  This demonstrates that the drone images in Set C have almost the same spatial resolution as the handheld images.  We believe going forward that the Merged Model will performed best on images captured by future drones, whose camera resolution will only continue to improve.

% Re-Annotated Validation Metrics for Full, Drone, Handheld Berry Models - Cropped Images, Set A
%\FloatBarrier
\setcounter{topnumber}{10}
\begin{table}[H]
\begin{center}
\begin{tabular}{|c|c|c|c|c|c|c|} 
\hline
Bush-Cropped & Prec. & Rec. &  Prec. & Rec. & Prec. & Rec. \\
Berry Model & (Green) & (Green) & (Blue) & (Blue) & (Overall)   & (Overall)    \\
\hline 
Drone & 0.778 &	0.675 &	0.733 &	0.168 &	0.776 &	0.631   \\
\hline 
Handheld & 0.483 &	0.007 &	0.069 &	0.019 &	0.222 &	0.008   \\
\hline
Merged & 0.758 &	0.712 &	0.811 &	0.265 &	0.76 &	0.673   \\
\hline
\end{tabular}
\caption{Bush-Cropped Berry Model (Drone, Handheld, Merged): Validation metrics for Validation Set A
\label{tab:re-val-metrics-cropped-set-a}}
\end{center}
\end{table}

% Re-Annotated Validation Metrics for Full, Drone, Handheld Berry Models - Cropped Images, Set B
%\FloatBarrier
\setcounter{topnumber}{10}
\begin{table}[H]
\begin{center}
\begin{tabular}{|c|c|c|c|c|c|c|} 
\hline
Bush-Cropped & Prec. & Rec. &  Prec. & Rec. & Prec. & Rec. \\
Berry Model & (Green) & (Green) & (Blue) & (Blue) & (Overall)   & (Overall)    \\
\hline 
Drone &  0.683 &	0.688 &	0.727 &	0.251 &	0.684 &	0.655  \\
\hline 
Handheld & 0.753 &	0.154 &	0.171 &	0.328 &	0.504 &	0.167   \\
\hline
Merged & 0.601 &	0.747 &	0.746 &	0.35 &	0.605 &	0.718   \\
\hline
\end{tabular}
\caption{Bush-Cropped Berry Model (Drone, Handheld, Merged): Validation metrics for Validation Set B
\label{tab:re-val-metrics-cropped-set-b}}
\end{center}
\end{table}

% Re-Annotated Validation Metrics for Full, Drone, Handheld Berry Models - Cropped Images, Set C
%\FloatBarrier
\setcounter{topnumber}{10}
\begin{table}[H]
\begin{center}
\begin{tabular}{|c|c|c|c|c|c|c|} 
\hline
Bush-Cropped & Prec. & Rec. &  Prec. & Rec. & Prec. & Rec. \\
Berry Model & (Green) & (Green) & (Blue) & (Blue) & (Overall)   & (Overall)    \\
\hline 
Drone & 0.708 &	0.583 &	0.848 &	0.618 &	0.752 &	0.595   \\
\hline 
Handheld & 0.799 &	0.589 &	0.837 &	0.615 &	0.812 &	0.598   \\
\hline
Merged & 0.809 &	0.617 &	0.847 &	0.605 &	0.821 &	0.613   \\
\hline
\end{tabular}
\caption{Bush-Cropped Berry Model (Drone, Handheld, Merged): Validation metrics for Validation Set C
\label{tab:re-val-metrics-cropped-set-c}}
\end{center}
\end{table}

For comparison, Table \ref{tab:re-val-metrics-bush-cropped-berry-sets-a-b-c} isolates for Validation Sets A, B, and C, all validated using the same Bush-Cropped Merged Berry Model.  Here, the results are similar to those when validated on entire images (see Table \ref{tab:re-val-metrics-merged-berry-model-fullsize-sets-a-b-c}), but observe that recall significantly improved for Set C due to the fact that the model no longer needs to detect tiny berries on background bushes, which it had difficulty with when validating on entire images.

% Validation Metrics for Combined Berry-Bush Model Cropped Bush, Sets A,B,C
%\FloatBarrier
\setcounter{topnumber}{10}
\begin{table}[htbp!]
\begin{center}
\begin{tabular}{|c|c|c|c|c|c|c|} 
\hline
Validation & Prec. & Rec. &  Prec. & Rec. & Prec. & Rec. \\
Dataset & (Green) & (Green) & (Blue) & (Blue) & (Overall)   & (Overall)   \\
\hline 
Set A & 0.758 &	0.712 &	0.811 &	0.265 &	0.76 &	0.673   \\
\hline 
Set B & 0.601 & 0.747 &	0.746 &	0.35 &	0.605 &	0.718  \\
\hline
Set C & 0.809 &	0.617 &	0.847 &	0.605 &	0.821 &	0.613  \\
\hline
\end{tabular}
\caption{Bush-Cropped Merged Berry Model: Validation metrics for Validation Sets 
A, B, C
\label{tab:re-val-metrics-bush-cropped-berry-sets-a-b-c}}
\end{center}
\end{table}

\subsection{Estimation of Picked-Visual Ratio}

The Bush-Cropped (Merged) Berry Model allows us to estimate the Picked-Visual Ratio (PVR) $\alpha$ (see Part 5 of Section III) by using its detections as an estimate for the Picked Ground Truth. 
We denote by $\alpha_p$ (predicted $\alpha$) to be any approximation of $\alpha$ calculated based on this estimate:
\begin{equation}
\alpha_p = \frac{\textrm{Picked GT}}{\textrm{Detections}} \approx \frac{\textrm{Picked GT}}{\textrm{Visual GT}} = \alpha
\end{equation}
On the other hand, we distinguish $\alpha_p$ from experimental values of $\alpha$ calculated by using the annotated visual GT of the cropped image, i.e., number of berry annotations within the central bounding box), which we assume to be a very accurate estimate of the true value of $\alpha$.

Tables \ref{tab:alpha-validation-set-a}, \ref{tab:alpha-validation-set-b}, and \ref{tab:alpha-validation-set-c} give both predicted and experimental values for $\alpha$ for Validation Sets A, B, and C, respectively, including he total number of detections, visual GT, and picked GT are given for each image (cropped around the foreground center bush).  Although the total number of detections appear to be good approximations of the visual GT for Validation Sets A and B (Tables \ref{tab:alpha-validation-set-a} and \ref{tab:alpha-validation-set-b}), this is misleading as these detections contain many false positives of berries (see overall precision and recall for the Bush-Cropped (Merge) Berry Model in Table \ref{tab:re-val-metrics-bush-cropped-berry-sets-a-b-c}), which cancel out the many false negatives of berries that were not detected. Thus, an accurate estimation of $\alpha_p$ will depend on an accurate Berry Model. 

Experimental values of $\alpha$ differ widely for all three sets (Tables \ref{tab:alpha-validation-set-a}, \ref{tab:alpha-validation-set-b}, and \ref{tab:alpha-validation-set-c}), with mean experimental values highest for Set $A$ and lowest for Set C.  This shows that estimating $\alpha$ will be challenging as it appears to depend not only on the blueberry variety (recall that Sets A and C correspond to Duke and Draper varieties, respectively), but also which side of the bush is captured (recall that Sets A and B correspond to two sides of the same five bushes).

%\FloatBarrier
% Calculation of alpha - Berry Validation Set A, Side 1
%\FloatBarrier
% List of images
% A1 - DJI_0002
% A2 - DJI_0003
% A3 - DJI_0004
% A4 - DJI_0005
% A5 - DJI_0006
\setcounter{topnumber}{10}
\begin{table}[htbp!]
\begin{center}
\begin{tabular}{|c|c|c|c|c|c|} 
\hline
Image &  Detections & Visual & Picked & $\alpha_p$ & $\alpha$ \\
 & & GT & GT & (Predicted) & (Experimental) \\
\hline 
A1 	&	882  &   1010 &   3312  & 3.755 & 3.279 \\
\hline 
A2 	&	1451 &   1230 &   3996  & 2.754 & 3.249 \\
\hline 
A3 	&	511  &   493  &   2888  & 5.652 & 5.858 \\
\hline 
A4 	&	711  &   847  &   2920  & 4.107 & 3.447 \\
\hline
A5 	&	420  &   708  &   1404  & 3.343 & 1.983 \\
\hline
\hline
%\verb|MEAN:| 
Mean      	    & 795 & 	858 &	2904 &	3.92 &	3.56     \\
\hline
%\verb|STANDARD DEVIATION:| 
SD        	    & 408 &	282 &	950 &	1.09 &	1.41    \\
\hline
\hline
%\verb|MEAN:| 
Total         & 3975 &	4288 &	14520 &	3.65 &	3.39   \\
\hline
\end{tabular}
\caption{Calculation of $\alpha$ (predicted vs experimental) for Berry Validation Set A using Bush-Cropped (Merged) Berry Model.}
\label{tab:alpha-validation-set-a}
\end{center}
\end{table}
%\FloatBarrier

%\FloatBarrier
% Calculation of alpha - Berry Validation Set B, Side 2
%\FloatBarrier
% List of images
% B1 - DJI_0018
% B2 - DJI_0019
% B3 - DJI_0020
% B4 - DJI_0021
% B5 - DJI_0022
\setcounter{topnumber}{10}
\begin{table}[htbp!]
\begin{center}
\begin{tabular}{|c|c|c|c|c|c|} 
\hline
Image &  Detections & Visual & Picked & $\alpha$ & $\alpha$ \\
 & & GT & GT & (Predicted) & (Experimental) \\
\hline
B1 	&	891  &   785  &   1404  & 1.576 & 1.789 \\
\hline
B2 	&	885  &   806  &   2920  & 3.299 & 3.623 \\
\hline
B3 	&	1012  &  972  &   2888  & 2.854 & 2.971 \\
\hline
B4 	&	1856  &  1842 &   3996  & 2.153 & 2.169 \\
\hline
B5 & 	1071  &  1043 &   3312  & 3.092 & 3.175 \\
\hline
\hline
%\verb|MEAN:| 
Mean      	    & 1143 &	1090 &	2904 &	2.59 &	2.75    \\
\hline
%\verb|STANDARD DEVIATION:| 
SD        	    & 406 &	435 &	950 &	0.71 &	0.75  \\
\hline
\hline
%\verb|MEAN:| 
Total         & 5715 &	5448 &	14520 &	2.54 &	2.67   \\
\hline
\end{tabular}
\caption{Calculation of $\alpha$ (predicted vs experimental) for Berry Validation Set B using Bush-Cropped (Merged) Berry Model.}
\label{tab:alpha-validation-set-b}
\end{center}
\end{table}
%\FloatBarrier

%\FloatBarrier
% Re-Annotated 2023 Validation Metrics - Model-Cropped Images, Validation Set C (Draper)
%\FloatBarrier
% List of images
% C1 - DJI_0465
% C2 - DJI_0466
% C3 - DJI_0468
% C4 - DJI_0470
% C5 - DJI_0474
\setcounter{topnumber}{10}
\begin{table}[htbp!]
\begin{center}
\begin{tabular}{|c|c|c|c|c|c|} 
\hline
Image & Detections & Visual & Picked & $\alpha$ & $\alpha$ \\
& & GT & GT & (Predicted) & (Experimental) \\
\hline 
C1 	&	1109  &   1507 &   2407 & 2.170  & 1.597\\
\hline 
C2 	&	831 &   924 &   3215  & 3.869 & 3.479\\
\hline 
C3 	&	1261  &   1491  &   1963 & 1.557  & 1.316\\
\hline 
C4 	&	713  &   618  &   2307 & 3.236  & 3.733\\
\hline
C5 	&	1210  &   1457  &   1963 & 1.622  & 1.347\\
\hline
\hline
%\verb|MEAN:| 
Mean      	&   955 &	1199 &	2371 &	2.67 &	2.29 \\
\hline
%\verb|STANDARD DEVIATION:| 
SD        	    & 225 &	406 &	513 &	1.10 &	1.21 \\
\hline
\hline
%\verb|MEAN:| 
Total          &   4774 &	5997 &	11855 &	2.48 &	1.98 \\
\hline
\end{tabular}
\caption{Calculation of $\alpha$ (predicted vs experimental) for Berry Validation Set C using Bush-Cropped (Merged) Berry Model.}
\label{tab:alpha-validation-set-c}
\end{center}
\end{table}
%\FloatBarrier

\subsection{Discussion}
% ---------------------
Results of our Berry Model highlight challenges with annotating berries and training our models.  Berries that are difficult to discern due to their small size, especially those on background bushes, can lead to subjective annotations and thus an ambiguous ground truth.  Many false positive detections could be argued as true detections of berries depending on one's visual acuity, but difficult to confirm with certainty because of their low resolution.   Moreover, occlusion of partially hidden berries, camouflage of green berries by leaves, and shaded berries makes for training an accurate Berry Model quite challenging.

Results of the Bush Model clearly show that detecting bushes is not necessarily an easier task than detecting berries. Obviously, a bush is considerably larger than a berry; however, the complicated branch structure of a bush, in addition its branches possibly overlapping with a neighboring bush, creates challenges in training an accurate bush model. 

Results of the Bush-Cropped Berry Model show the effectiveness of cropping around the foreground center bush to eliminate background berries and thus improved the model's precision and recall, which in turn provided a more accurate estimation of crop yield.  Estimates of the Picked-Visual Ratio $\alpha$  based on the Bush-Cropped Berry Model show that it can variety significantly and depends on many factors such as the particular side of the bush that is captured and the blueberry variety, and other factors that we did not take into account: bush size, bush foliage density, environmental and soil conditions.

\section{Conclusion}
% -----------------------------------------------
In this paper we presented a pipeline of object detection models based on deep learning for detecting blueberry bushes and individual berries on them.  These models allow a smart drone programmed with them to fly intelligent missions, namely to precisely locate bushes and capture their side views, thus obtaining a more accurate estimate of crop yield.  We have already begin to test our pipeline using a custom-build programmable drone to capture data and hope to report on our experimental results in the near future.  
We hope our work will spur interest in others to address the challenges raised in this paper and improve on our baseline results.  All datasets, models, and source code will be made available on Github.

%\subsection{Acknowledgements}
% ---------------------

\section*{Acknowledgment}

The authors would like to acknowledge partial financial support from 
%Rowan University and 
the New Jersey Council of County Colleges through their NJ Pathways to Career Opportunities Program, Department of Mathematics and College of Science and Mathematics at Rowan University. The following blueberry farms in South Jersey kindly provided us access to their fields in order to collect data: Macrie Brothers Farm, Moore's Meadow Farm, and Vacarrella Farm.  We also wish to thank other former and current team members who contributed to annotating our datasets: 
Robert Czarnota,
Jacob Green,
Lori Green,
Felix Hakimi,
Lance Ilagan,
Nicholas Kaegi,
Jamie Kahle,
Brian Kim,
Tuan Le, 
Ik Jae Lee,
Duy Nguyen,
Jonah Rodriguez, and
Iosefa Sunia.

%\pagebreak
%\clearpage
% \section*{References}

\bibliographystyle{IEEEtran}
\bibliography{blueberry_crop_yield}

\begin{thebibliography}{10}
\providecommand{\url}[1]{#1}
\csname url@samestyle\endcsname
\providecommand{\newblock}{\relax}
\providecommand{\bibinfo}[2]{#2}
\providecommand{\BIBentrySTDinterwordspacing}{\spaceskip=0pt\relax}
\providecommand{\BIBentryALTinterwordstretchfactor}{4}
\providecommand{\BIBentryALTinterwordspacing}{\spaceskip=\fontdimen2\font plus
\BIBentryALTinterwordstretchfactor\fontdimen3\font minus \fontdimen4\font\relax}
\providecommand{\BIBforeignlanguage}[2]{{%
\expandafter\ifx\csname l@#1\endcsname\relax
\typeout{** WARNING: IEEEtran.bst: No hyphenation pattern has been}%
\typeout{** loaded for the language `#1'. Using the pattern for}%
\typeout{** the default language instead.}%
\else
\language=\csname l@#1\endcsname
\fi
#2}}
\providecommand{\BIBdecl}{\relax}
\BIBdecl

\bibitem{Redmon_2016_CVPR}
J.~Redmon, S.~Divvala, R.~Girshick, and A.~Farhadi, ``You only look once: Unified, real-time object detection,'' in \emph{Proceedings of the IEEE Conference on Computer Vision and Pattern Recognition (CVPR)}, June 2016.

\bibitem{cropyield-survey}
T.~van Klompenburg, A.~Kassahun, and C.~Catal, ``Crop yield prediction using machine learning: A systematic literature review,'' \emph{Computers and Electronics in Agriculture}, vol. 177, pp. 1--18, 2020.

\bibitem{apple-faster-rcnn}
S.~Yildirim, , and B.~Ulu, ``Deep learning based apples counting for yield forecast using proposed flying robotic system,'' \emph{Sensors}, vol.~23, pp. 1--14, 2023.

\bibitem{apple-yolo-improved}
H.~Wang, J.~Feng, and H.~Yin, ``Improved method for apple fruit target detection based on yolov5s,'' \emph{Agriculture}, vol.~13, pp. 1--26, 2023.

\bibitem{intelligent-fruit-detection}
O.~Melnychenko, L.~Scislo, O.~Savenko, A.~Sachenko, and P.~Radiuk, ``Intelligent integrated system for fruit detection using multi-uav imaging and deep learning,'' \emph{Sensors}, vol.~24, no.~6, pp. 1--36, 2024.

\bibitem{apple-yield-mapping}
N.~Hani, P.~Roy, and V.~Isler, ``A comparative study of fruit detection and counting methods for yield mapping in apple orchards,'' \emph{Journal of Field Robotics}, vol.~37, pp. 263--282, 2020.

\bibitem{drone-tomato}
Y.~Egi, M.~Hajyzadeh, and E.~Eyceyurt, ``Drone-computer communication based tomato generative organ counting model using yolo v5 and deep-sort,'' \emph{Agriculture}, vol.~12, no.~9, pp. 1--17, 2022.

\bibitem{yield-precision-ag}
Y.~Osman, R.~Dennis, and K.~Elgazzar, ``Yield estimation and visualization solution for precision agriculture,'' \emph{Sensors}, vol.~21, pp. 1--25, 2021.

\bibitem{treetop}
P.~Hofinger, H.-J. Klemmt, S.~Ecke, S.~Rogg, and J.~Dempewolf, ``Application of yolov5 for point label based object detection of black pine trees with vitality losses in uav data,'' \emph{Remote Sens.}, vol.~15, pp. 1--13, 2023.

\bibitem{cranberry}
P.~Akiva, K.~Dana, P.~Oudemans, and M.~Mars, ``Finding berries: Segmentation and counting of cranberries using point supervision and shape priors,'' in \emph{2020 IEEE/CVF Conference on Computer Vision and Pattern Recognition Workshops (CVPRW)}, 2020, pp. 219--228.

\bibitem{grape-cluster-yolo}
L.~Shen, J.~Su, R.~He, L.~Song, R.~Huang, Y.~Fang, Y.~Song, and B.~Su, ``Real-time tracking and counting of grape clusters in the field based on channel pruning with yolov5s,'' \emph{Computers and Electronics in Agriculture}, vol. 206, pp. 1--14, 2023.

\bibitem{grape-bunch-damage-yolo}
I.~Pinheiro, G.~Moreira, D.~Queirós~da Silva, S.~Magalh\~{a}es, A.~Valente, P.~M. Oliveira, M.~Cunha, and F.~Santos, ``Deep learning yolo-based solution for grape bunch detection and assessment of biophysical lesions,'' \emph{Computers and Electronics in Agriculture}, vol. 206, pp. 1--14, 2023.

\bibitem{blueberry-maturity-yield}
C.~B. MacEachern, T.~J. Esau, A.~W. Schumann, P.~J. Hennessy, and Q.~U. Zaman, ``Detection of fruit maturity stage and yield estimation in wild blueberry using deep learning convolutional neural networks,'' \emph{Smart Agricultural Technology}, vol.~3, pp. 1--11, 2023.

\bibitem{blueberry-ripeness}
W.~Yang, X.~Ma, and H.~An, ``Blueberry ripeness detection model based on enhanced detail feature and content-aware reassembly,'' \emph{Agronomy}, vol.~13, pp. 1--19, 2023.

\bibitem{blueberry-traits-segmentation}
X.~Ni, C.~Li, H.~Jiang, and F.~Takeda, ``Deep learning image segmentation and extraction of blueberry fruit traits associated with harvestability and yield,'' \emph{Horticulture Research}, vol.~7, no.~1, pp. 1--14, 2020.

\bibitem{blueberry-row-segmentation}
D.~Stefanović, A.~Antić, M.~Otlokan, B.~Ivošević, O.~Marko, V.~Crnojević, and M.~Panić, ``Blueberry row detection based on uav images for inferring the allowed ugv path in the field,'' in \emph{ROBOT2022: Fifth Iberian Robotics Conference.}, 2022.

\bibitem{bush-detection}
V.~Filipovi\'{c}, D.~Stefanovi\'{c}, N.~Pajevi\'{c}, Z.~Grbovi\'{c}, N.~Djuric, and M.~Pani\'{c}, ``Bush detection for vision-based ugv guidance in blueberry orchards: Data set and methods,'' in \emph{2023 IEEE/CVF Conference on Computer Vision and Pattern Recognition Workshops (CVPRW)}, 2023, pp. 3646--3655.

\bibitem{apple-fruit-actual-count}
S.~Bargoti and J.~Underwood, ``Image segmentation for fruit detection and yield estimation in apple orchards,'' \emph{J. Field Robot.}, vol.~34, no.~6, pp. 1039--1060, 2016.

\bibitem{mango-fruit-actual-count}
M.~Stein, S.~Bargoti, and J.~Underwood, ``Image based mango fruit detection, localisation and yield estimation using multiple view geometry,'' \emph{Sensors}, vol.~16, no.~11, 2016.

\bibitem{mango-fruit-segmentation-actual-count}
A.~B. Payne, K.~B. Walsh, P.~Subedi, and D.~Jarvis, ``Estimation of mango crop yield using image analysis – segmentation methodn, localisation and yield estimation using multiple view geometry,'' \emph{Computers and Electronics in Agriculture}, vol.~91, pp. 57--64, 2013.

\bibitem{almond-fruit-count}
J.~Underwood, C.~Hung, B.~Whelan, and S.~Sukkarieh, ``Mapping almond orchard canopy volume, flowers, fruit and yield using lidar and vision sensors,'' \emph{Computers and Electronics in Agriculture}, vol. 130, pp. 83--96, 2016.

\bibitem{grape-fruit-count}
F.~Palacios, M.~P. Diago, P.~Melo-Pinto, and J.~Tardaguila, ``Early yield prediction in different grapevine varieties using computer vision and machine learning,'' \emph{Precision agriculture}, vol.~24, no.~2, pp. 407--435, 2023.

\bibitem{deepsort}
N.~Wojke, A.~Bewley, and D.~Paulus, ``Simple online and realtime tracking with a deep association metric,'' in \emph{2017 IEEE International Conference on Image Processing (ICIP)}, 2017, pp. 3645--3649.

\end{thebibliography}

\end{document}